\documentclass{article}

\usepackage{arxiv}

\usepackage[utf8]{inputenc} 
\usepackage[T1]{fontenc}    
\usepackage{hyperref}       
\usepackage{url}            
\usepackage{booktabs}       
\usepackage{amsfonts}       
\usepackage{nicefrac}       
\usepackage{microtype}      
\usepackage{lipsum}		
\usepackage{graphicx}
\usepackage{natbib}
\usepackage{doi}
\usepackage{amsmath}
\usepackage{multirow}
\usepackage{amssymb}
\usepackage{bm}

\title{Let the Data Decide: Supervision Analysis, Capability Trade-offs, and Adaptive Objective Routing in Continued Pre-Training via Off-Policy Distillation}

\author{
Jiangan Yuan\thanks{Co-first authors with equal contributions.} \\
Baidu Inc. \\
\texttt{jianganyuan@link.cuhk.edu.cn}
\And
Zhixuan Li\makeatletter\footnotemark[1]\makeatother \\
Baidu Inc. \\
\texttt{lizhixuan.2017@tsinghua.org.cn}
\And
Han Xu \\
Baidu Inc. \\
\texttt{xhbj66@gmail.com}
}

\renewcommand{\shorttitle}{Let the Data Decide}

\hypersetup{
pdftitle={A template for the arxiv style},
pdfsubject={q-bio.NC, q-bio.QM},
pdfauthor={David S.~Hippocampus, Elias D.~Striatum},
pdfkeywords={First keyword, Second keyword, More},
}

\begin{document}
\maketitle

\begin{abstract}
Off-policy distillation is now central to large language model pre-training, yet how training data, objective parameterization, and model capabilities interact remains poorly characterized. We studies top-$k$-truncated, temperature-scaled off-policy distillation by decomposing this problem into two questions: an \emph{objective-to-capability} analysis of how the training objective shapes token-level supervision and downstream performance, and a \emph{data-to-objective} analysis of how data heterogeneity should inform objective routing. We first show that the language-modeling objective ($L_{\mathrm{LM}}$) and the knowledge-distillation objective ($L_{\mathrm{KD}}$) induce systematically different capability profiles, and trace this divergence to a gradient-level tension between \emph{direct observed-token reinforcement} and \emph{teacher-supported alternative supervision}. To quantify this tension, we introduce diagnostic metrics---support coverage, observed-token probability mass, and teacher-distribution concentration---and show via controlled sweeps that the support size $k$ governs a coverage--sharpness trade-off, while distillation temperature controls within-support probability allocation. We then examine adaptive objective routing: a domain-level policy that applies $L_{\mathrm{LM}}$ to math and code and $L_{\mathrm{KD}}$ to general-domain data yields consistent gains over both single-objective baselines, whereas token-level routing based on observed-token probability mass or teacher entropy fails to consistently match the single-objective baseline. These results suggest that effective objective routing depends less on routing granularity than on the quality of the routing signal, reframing continued pre-training via off-policy distillation as a structured, data-conditional supervision-design problem rather than a global hyperparameter choice.
\end{abstract}

\section{Introduction}

Off-policy distillation has become an increasingly important ingredient in large language model (LLM) pre-training. In this setting, a student model is trained on a fixed corpus while receiving supplementary supervision from a larger, frozen teacher model, typically in the form of soft next-token distributions over the vocabulary. Compared with standard language-modeling (LM) training on one-hot targets, knowledge distillation (KD) can expose the student to the teacher's distributional uncertainty over plausible continuations---an effect commonly attributed to the ``dark knowledge'' encoded in soft probability distributions~\citep{hinton2015distilling}---rather than grounding training exclusively on the single observed token. These properties make off-policy distillation attractive for improving data efficiency, transferring capabilities from stronger models, and recovering performance after structural compression. Recent industrial model families have demonstrated that distillation can serve not merely as a post-training compression tool but as a central mechanism for constructing smaller and more capable base models~\citep{team2024gemma2,team2025gemma3,muralidharan2024compact,liu2026ministral,tang2026slimqwen}.

Despite its practical importance, off-policy distillation remains insufficiently characterized as a training objective. Existing work has produced a range of useful empirical recipes, but two broader limitations persist. First, most analyses treat data composition, objective choice, sparse teacher-target construction, and downstream benchmark behavior as separate design dimensions, even though they jointly determine the token-level supervision signal that the student receives at each training position. This fragmentation makes it difficult to reason systematically about how the training objective interacts with data heterogeneity to shape downstream model capabilities. Second, many reported findings remain primarily phenomenological: they document that adjusting a loss weight, a top-$k$ cutoff, a temperature parameter, or a routing heuristic changes benchmark scores, but offer limited explanation of which token-level properties of the supervision signal are responsible or why the effect takes the observed form. As a result, off-policy distillation is often treated as a global recipe to be tuned empirically rather than as a structured, data-dependent training objective whose effects can be systematically analyzed and anticipated.

These limitations manifest in three concrete ways. The first concerns the relationship between training objective and model capability. Most distillation pipelines apply a fixed interpolation between LM and KD losses uniformly across all token positions. Some work reports that pure KD can outperform LM--KD mixtures in certain settings~\citep{muralidharan2024compact,liu2026ministral}, while others study static or dynamic weighting schedules~\citep{peng2025pre}. These findings establish that the LM--KD balance is a consequential design choice, but evaluations are typically reported as aggregate benchmark scores. They do not identify which capability dimensions each objective preferentially strengthens, whether benchmark performance diverges systematically across task types under fixed global objectives, or whether a single global objective necessarily incurs an avoidable capability trade-off across heterogeneous tasks.

The second limitation concerns the parameterization of the KD objective itself. Modern LLM vocabularies make full-vocabulary teacher logits prohibitively expensive to store and transmit at scale, so practical systems typically construct sparse teacher targets via top-$k$ truncation or sampling, followed by temperature scaling and renormalization~\citep{team2025gemma3,peng2025pre,goyal2026distilled}. These choices directly determine the support, sharpness, and probability allocation of the target distribution, and therefore materially affect the optimization signal that the student receives at each position. Existing work has explored sweeps over $k$ or temperature and reported the resulting downstream benchmark scores, but without a detailed account of how each parameter reshapes the token-level supervision landscape or why its downstream effect takes the observed form.

The third limitation concerns the relationship between data heterogeneity and objective routing. Pre-training corpora are intrinsically heterogeneous, and different domains, examples, or individual token positions may benefit from different training objectives. Recent work has begun to explore token-level routing policies, using teacher entropy, distribution sharpness, or related difficulty signals to determine when to apply KD and when to default to standard LM supervision~\citep{peng2025pre,goyal2026distilled,zhang2026egad,xie2026llm,huang2026selectkd}. However, the design of routing signals is rarely analyzed in depth. In particular, it remains unclear whether teacher-side statistics such as entropy are sufficient for reliable routing, or whether effective objective routing requires a joint measure of alignment between the teacher distribution and the observed training token.

This paper studies continued pre-training via off-policy distillation through the interplay of data heterogeneity, training objective, and downstream capability. We decompose the problem into two questions. The first is an \emph{objective-to-capability} question: given the same training data and model initialization, how do LM and KD supervision differ at the token level, and how do KD-specific parameters such as top-$k$ and temperature reshape the resulting capability profile? The second is a \emph{data-to-objective} question: given a heterogeneous training corpus, can different data subsets or individual token positions be routed to different objectives to better align supervision with the underlying training distribution, and what properties make a routing signal reliable? Together, these questions reframe continued pre-training via off-policy distillation not as a monolithic hyperparameter choice but as a problem of structured, data-conditional supervision design.

We address both questions through a combination of gradient-level analysis, diagnostic measurement, controlled hyperparameter sweeps, and objective routing experiments. We begin by directly comparing a standard LM-trained student with an off-policy KD-trained student under otherwise identical training conditions. The results reveal that neither objective uniformly dominates: the two objectives induce systematically distinct capability profiles. LM supervision is consistently stronger on high-difficulty reasoning, mathematical problem solving, and knowledge-intensive evaluations, with the performance gap amplified substantially under Pass@$K$ evaluation at larger sampling budgets. KD supervision is more favorable on benchmarks associated with commonsense plausibility, factual retrieval, local reading comprehension, and structured program synthesis. This systematic, benchmark-dependent divergence demonstrates that a fixed global objective commits to a particular capability trade-off rather than being uniformly optimal across heterogeneous tasks, directly motivating a principled analysis of where and why the two objectives differ.

To explain this divergence, we analyze the two objectives at the gradient level. Standard LM training concentrates all supervision on the observed token, while top-$k$-truncated KD distributes target mass across the teacher support, either reinforcing or suppressing the observed-token update depending on whether and to what degree the teacher assigns probability mass to that token. This analysis identifies two complementary training signals---\emph{direct observed-token reinforcement} and \emph{teacher-supported alternative supervision}---whose relative balance governs the capability trade-off between the objectives. To quantify this balance at each training position, we introduce a suite of diagnostic metrics: observed-token coverage by the teacher support, observed-token rank, observed-token mass (OTMass), conditional observed-token mass (CondOTMass), and teacher-support entropy. These metrics characterize how the sparse teacher target departs from the hard LM target along each dimension of the supervision signal and serve as instruments for connecting token-level optimization statistics to downstream benchmark behavior.

Using these diagnostics, we conduct controlled sweeps over the two principal degrees of freedom in sparse KD: support size $k$ and distillation temperature $\tau$. The support size governs a coverage--sharpness trade-off: small supports produce sharp, concentrated supervision but frequently exclude the observed token, while large supports reduce coverage gaps at the cost of a progressively more diffuse target. Temperature, by contrast, does not alter support membership; at fixed $k$, it controls within-support probability allocation, sharpening the target toward the teacher's top-ranked prediction at low temperatures and redistributing mass toward lower-ranked alternatives at high temperatures. Across standard benchmarks and Pass@$K$ evaluations, the effects of these two parameters are qualitatively distinct and task-dependent, consistent with the interpretation that $k$ and $\tau$ jointly position the supervision signal along the spectrum between the two complementary factors identified above, and that different tasks favor different positions along this spectrum.

We then investigate whether objective routing can be made adaptive to the training data. At the coarsest granularity, a domain-based routing policy that applies LM supervision to mathematics and code data and KD supervision to general-domain data yields consistent aggregate improvements over both single-objective baselines: it largely recovers the LM advantage on reasoning-intensive benchmarks and Pass@$K$ evaluations, preserves the KD advantage on knowledge-oriented and commonsense benchmarks, and in several cases---including MBPP and the AIME benchmarks---exceeds both baselines simultaneously. Analysis of the diagnostic metrics stratified by domain provides a mechanistic account of this behavior: mathematics and code data exhibit substantially higher observed-token mass and lower teacher-support entropy than general-domain data, indicating that domain labels serve as a reliable proxy for teacher--data alignment rather than merely indexing surface-level content categories.

Finally, we compare domain-level routing with finer-grained token-level routing policies that use OTMass and teacher entropy directly as per-token routing signals. Despite operating at finer granularity, neither token-level policy consistently matches the stronger single-objective baseline, and both fall substantially short of domain-level routing on high-difficulty reasoning evaluations. Between the two token-level signals, OTMass is the more reliable routing criterion: it directly measures the teacher probability mass assigned to the observed training token and thereby captures teacher--data alignment, whereas teacher entropy characterizes only the concentration of the teacher distribution independently of the observed token and cannot distinguish a teacher that agrees with the data from one that confidently predicts an alternative. These results suggest that routing effectiveness need not increase monotonically with granularity, and that reliable objective routing may instead call for signals that jointly reflect the teacher distribution and its alignment with the observed training token---a criterion that teacher-side uncertainty measures alone do not appear to satisfy.
\section{Related Work}

\subsection{Off-policy Distillation in Pre-training}

Knowledge distillation has long been used to compress pretrained language models, and several early works already operate in an off-policy manner by training a student on a fixed corpus with a frozen teacher. DistilBERT introduces a pre-training recipe that combines masked-language modeling, soft-target distillation, and cosine representation matching, demonstrating that distillation can be applied during pre-training rather than solely at downstream fine-tuning time~\citep{sanh2019distilbert}. TinyBERT extends this approach with a two-stage framework that distills both general-domain pre-training knowledge and task-specific knowledge through Transformer-layer supervision~\citep{jiao2020tinybert}. MobileBERT, MiniLM, and MiniLMv2 further show that architecture-aware distillation and self-attention relation transfer can preserve much of the teacher quality with fewer layers and lower latency~\citep{sun2020mobilebert,wang2020minilm,wang2021minilmv2}. Although these works focus primarily on encoder-style PLMs rather than decoder-only LLMs, they establish the foundational template of offline teacher--student transfer on a fixed corpus.

More recent work has revisited this paradigm at the scale of modern LLM pre-training. The Gemma series offers a prominent industrial-scale demonstration. Gemma~2 trains its 2B and 9B models with knowledge distillation, motivating this choice as a richer training signal capable of providing denser gradients than the observed token alone~\citep{team2024gemma2}. Gemma~3 further extends this recipe to all released model sizes and reports a practical sparse logit-sampling implementation for efficient teacher-target construction~\citep{team2025gemma3}.

A related line of work combines structured pruning with distillation to produce model families spanning multiple deployment scales. Minitron demonstrates that a pretrained dense model can be pruned and retrained with knowledge distillation using only a small fraction of the original training data, and its empirical study emphasizes that distillation is critical for performance recovery after pruning~\citep{muralidharan2024compact}. Ministral~3 follows a similar compression-oriented direction through cascade distillation~\citep{liu2026ministral}, and SlimQwen extends this theme to MoE compression~\citep{tang2026slimqwen}. These works position distillation not only as a standalone pre-training objective but also as a recovery mechanism that makes aggressive architecture compression feasible.

Beyond industrial recipes, recent academic studies have begun to analyze the design space and behavioral consequences of pre-training distillation. \citet{peng2025pre} systematically investigate top-$p$--top-$k$ truncation for reducing logit storage, temperature normalization of teacher distributions, static and scheduled mixtures of LM and KD losses, and adaptive temperature variants. \citet{goyal2026distilled} offer a complementary mechanistic perspective by analyzing distilled pre-training through the lens of in-context learning and test-time scaling, showing that distillation can improve test-time scaling by increasing generation diversity while simultaneously impairing in-context learning at low-entropy positions. Together, these works motivate a closer examination of how off-policy distillation reshapes token-level supervision during pre-training.

\subsection{KL Objectives for Off-policy Distillation}

The standard logits-based distillation objective traces back to the soft-target formulation of \citet{hinton2015distilling}. The student is trained to match the softened teacher distribution, typically in combination with a hard-label cross-entropy term. In token-level language-model pre-training, this idea is most commonly instantiated as a forward KL loss over next-token distributions. Let $p_T(\cdot \mid x_{<r})$ and $p_S(\cdot \mid x_{<r})$ denote the teacher and student distributions, respectively. A generic off-policy distillation objective can be written as
\[
L
=
(1-\alpha)L_{\mathrm{LM}}
+
\alpha L_{\mathrm{KD}},
\qquad
L_{\mathrm{KD}}
=
\frac{1}{T}\sum_{t=1}^{T}
D_{\mathrm{KL}}\!\left(
\tilde p_T(\cdot \mid x_{<r})
\,\|\, 
p_S(\cdot \mid x_{<r})
\right),
\]
where $\tilde p_T$ denotes the processed teacher target distribution after operations such as sampling, scaling and renormalization.

Most large-scale pre-training distillation recipes adopt forward KL as the default token-level objective, differing primarily in how it is combined with the LM loss. \citet{peng2025pre} explicitly formulate pre-training distillation as a weighted combination of $L_{\mathrm{LM}}$ and $L_{\mathrm{KD}}$, and explore schedules for the mixing coefficient $\alpha$. \citet{goyal2026distilled} adopt a similar weighted objective to study distilled pre-training under modern LLM scaling conditions. By contrast, several pruning-and-distillation recipes find that pure logits distillation can be preferable following model compression~\citep{muralidharan2024compact,liu2026ministral}. These findings suggest that the LM--KD mixing ratio is not merely an implementation detail but a substantive design choice whose optimal value depends on the training regime, model initialization, and teacher--student capacity gap.

Although forward KL is the dominant choice in off-policy pre-training, several works study alternative divergence formulations in adjacent LLM distillation settings. MiniLLM argues that forward KL can be suboptimal for open-ended generation because it may force a capacity-limited student to cover too many teacher modes, and instead optimizes reverse KL with a policy-gradient-style procedure~\citep{gu2024minillm}. TAID takes a different approach by constructing an adaptive intermediate distribution between the student and teacher that is progressively interpolated over the course of training~\citep{shing2025taid}. These methods are developed primarily for post-training or general LLM distillation rather than the pre-training setting studied here; we therefore focus on the standard forward KL objective while treating these works as evidence that the choice and parameterization of the distillation target can materially affect student behavior.

Full-vocabulary teacher logits are expensive to store and transfer in practice since modern LLM vocabularies often contain hundreds of thousands of tokens. Off-policy distillation consequently relies on sparse approximations of the teacher distribution. One line of work uses deterministic truncation: \citet{peng2025pre} apply a top-$p$-then-top-$k$ truncation strategy before renormalizing the retained distribution. LFM2 introduces a top-$k$ teacher support with a decoupled objective that separates a binary KL term, which matches the total probability mass assigned to the teacher top-$k$ set, from a conditional KL term, which matches relative probabilities within that set~\citep{amini2025lfm2}. A complementary line of work samples sparse teacher targets rather than selecting the deterministic top-$k$ set. Gemma~3 samples 256 logits per token weighted by teacher probabilities, assigns zero probability to non-sampled entries, and renormalizes the resulting distribution~\citep{team2025gemma3}. \citet{goyal2026distilled} also study sparse-label distillation in which $k$ logits are sampled from the teacher distribution and renormalized. \citet{anshumann2025sparse} offer a more principled critique of deterministic top-$k$ distillation, arguing that it yields a biased estimate of the teacher distribution by discarding tail information, and propose random sampling with importance weighting. \citet{dasgupta2026don} further argue that standard KL is dominated by the teacher's highest-probability modes, and propose a tail-aware divergence that decouples top-$k$ and tail contributions.

Temperature is another central factor governing the shape of the distillation target. In the classical formulation of \citet{hinton2015distilling}, the same temperature is applied to both teacher and student logits, and the soft-target loss is multiplied by $\tau^2$ to compensate for the temperature-dependent gradient scale. In LLM pre-training distillation, however, temperature is typically employed as part of teacher-target construction: teacher logits are rescaled, then truncated or sampled, renormalized, and stored as sparse soft labels. \citet{goyal2026distilled} apply temperature to teacher logits when constructing sparse labels, and \citet{peng2025pre} systematically study both static and adaptive teacher-temperature variants based on distribution sharpness. Together, these works indicate that sparse support size and temperature are two coupled but distinct degrees of freedom in the design of off-policy distillation targets.

\subsection{Training Objective Routing}

Most distillation methods apply the training objective uniformly across all token positions. The standard formulation interpolates the hard-label language-modeling objective and the distillation objective as
\[
    L = (1-\alpha)L_{\mathrm{LM}} + \alpha L_{\mathrm{KD}},
\]
where $\alpha$ is applied uniformly at every token position.

Prior work has primarily adjusted the global mixture ratio or the training-stage schedule. \citet{peng2025pre} explore scheduled weighting between LM and KD losses in pre-training distillation, treating the balance between hard-label and teacher supervision as a time-dependent design choice. TAID achieves a form of temporal routing by replacing direct teacher matching with an adaptive intermediate distribution that evolves over the course of training~\citep{shing2025taid}. These approaches move beyond a fixed global objective but still assign the same objective form to every token within a training step or stage.

A finer-grained line of work performs token-conditional routing. \citet{goyal2026distilled} observe that distillation can harm in-context learning by softening low-entropy, near-deterministic token mappings, and propose entropy-based routing that removes the distillation term from the lowest-entropy positions while retaining standard hard-label supervision there. \citet{peng2025pre} also examine adaptive temperature rules based on distributional sharpness---measured by standard deviation or entropy---thereby treating the softness of the distillation target as a token-dependent quantity. \citet{zhang2026egad} use teacher entropy as a token-level difficulty signal and jointly adapt three components of distillation: the KL weight, the temperature, and the distillation path; their method emphasizes low-entropy tokens early in training and gradually shifts supervision toward high-entropy tokens, while also assigning token-specific temperatures and applying deeper feature and attention distillation to difficult tokens. \citet{xie2026llm} instead measure token difficulty by the Hellinger distance between teacher and student distributions: their LATF module dynamically selects tokens for the distillation loss, while their IDTS module assigns lower temperatures to difficult tokens and higher temperatures to easier ones. \citet{huang2026selectkd} further reframe distillation as a token-acceptance problem in which the student proposes tokens, the teacher verifies them through top-$k$ or Spec-$k$ acceptance, and only accepted tokens receive the full distillation loss while rejected tokens are masked or down-weighted. Collectively, these studies suggest that objective routing is becoming a central degree of freedom in distillation design, alongside sparse support construction and temperature scaling. However, the majority of fine-grained routing methods have been developed and evaluated in post-training or instruction-distillation settings; fine-grained objective routing during large-scale off-policy pre-training remains comparatively underexplored, leaving open how token- or subset-level objective choices should be made when the training corpus, teacher targets, and downstream capabilities are all heterogeneous.

\section{Preliminaries}
\label{sec:preliminaries}

Consider a training sample represented as a token sequence:
\begin{equation}
x = (x_1, x_2, \dots, x_n).
\end{equation}
Under autoregressive language modeling, the model predicts each target token $x_r$ conditioned on the preceding context $x_{<r}$. We consider a large teacher model $f_T$ and a smaller student model $f_S$, both evaluated on the same sequence. For $M \in \{T, S\}$, model $f_M$ produces a logit vector over the vocabulary $\mathcal{V}$ at each position $r$:
\begin{equation}
z_r^{(M)} = f_M(x_{<r}) \in \mathbb{R}^{|\mathcal{V}|}, \quad M \in \{T, S\}.
\end{equation}
Here, $z_{r,i}^{(M)}$ denotes the logit assigned by model $M$ to token $i \in \mathcal{V}$ at position $r$.

\subsection{Top-$k$ Truncation and Temperature-Scaled Teacher Probabilities}

Storing and transferring full teacher logit vectors incurs substantial overhead due to the large vocabulary size. To mitigate this, we apply top-$k$ truncation to the teacher outputs. Specifically, at each position $r$, we define the teacher top-$k$ support as
\begin{equation}
K_r^{(T)} = \operatorname{TopK}(z_r^{(T)}, k),
\end{equation}
where $K_r^{(T)} \subseteq \mathcal{V}$ is the index set of the $k$ largest teacher logits at position $r$.

The teacher logits are then restricted to this support:
\begin{equation}
\hat{z}_{r,i}^{(T)} =
\begin{cases}
z_{r,i}^{(T)}, & i \in K_r^{(T)}, \\[4pt]
-\infty, & i \notin K_r^{(T)}.
\end{cases}
\end{equation}

Following truncation, we optionally apply a temperature $\tau > 0$ to control the sharpness of the teacher distribution. The resulting teacher probability is
\begin{equation}
p_{r,i}^{(T,k,\tau)}
=
\begin{cases}
\displaystyle
\frac{\exp(z_{r,i}^{(T)} / \tau)}
{\sum_{j \in K_r^{(T)}} \exp(z_{r,j}^{(T)} / \tau)}, & i \in K_r^{(T)}, \\[12pt]
0, & i \notin K_r^{(T)}.
\end{cases}
\end{equation}

By contrast, student logits are generated and consumed on-the-fly during training, so they impose negligible storage or communication overhead. To retain as much optimization signal as possible, we compute the student distribution over the full vocabulary without top-$k$ truncation. Consistent with standard practice in off-policy distillation, we also omit temperature scaling for the student. The student probability is thus defined as
\begin{equation}
p_{r,i}^{(S)}
=
\frac{\exp(z_{r,i}^{(S)})}
{\sum_{j \in \mathcal{V}} \exp(z_{r,j}^{(S)})}.
\end{equation}

\subsection{Training Objectives}

To align the student distribution with the teacher's truncated soft targets, we employ the forward KL divergence. At the sequence level, the knowledge distillation loss is defined as
\begin{equation}
L_{\mathrm{KD}}
=
\frac{1}{n} \sum_{r=1}^{n}
D_{\mathrm{KL}}
\!\left(
p_r^{(T,k,\tau)}
\;\middle\|\;
p_r^{(S)}
\right)
=
\frac{1}{n} \sum_{r=1}^{n}
\sum_{i \in K_r^{(T)}}
p_{r,i}^{(T,k,\tau)}
\log
\frac{
p_{r,i}^{(T,k,\tau)}
}{
p_{r,i}^{(S)}
}.
\end{equation}
The summation is restricted to $K_r^{(T)}$ because $p_{r,i}^{(T,k,\tau)} = 0$ for all $i \notin K_r^{(T)}$. The student probability $p_{r,i}^{(S)}$ in the denominator is always computed over the full vocabulary.

Apart from the distillation objective, the student can also be trained with the standard language modeling (LM) loss on ground-truth tokens:
\begin{equation}
L_{\mathrm{LM}}
=
-\frac{1}{n} \sum_{r=1}^{n}
\log p_{r,x_r}^{(S)}.
\end{equation}
\section{Experimental Setup}
\label{sec:experimental_setup}
This section describes the experimental configuration shared across all analyses, including the model architecture, training data, and evaluation protocol.

\subsection{Model Setup}

\subsubsection{Pruning--Distillation Scaling Ladder}
Conducting controlled studies directly on a large-scale model is computationally prohibitive. At the same time, using an independently trained small model introduces potential confounds, as its behavior may differ from that of the large model for reasons unrelated to the training objective under investigation.

To reduce experimental cost while preserving a principled connection to the large-scale model, we construct the student model as the final rung of a multi-stage pruning--distillation scaling ladder. Specifically, we begin with DeepSeek V3 Base, denoted $M_0$, which has 671B total parameters and 37B activated parameters. We prune $M_0$ to 128B total parameters and 19B activated parameters, then apply distillation-based continued pre-training on 1T tokens to recover and stabilize performance; the resulting model is denoted $M_1$. This procedure is applied recursively: $M_1$ is pruned to 59B total parameters and 10B activated parameters, followed by 200B tokens of distillation-based continued pre-training to obtain $M_2$; $M_2$ is further pruned to 24B total parameters and 4.5B activated parameters, followed by an additional 200B tokens of distillation-based continued pre-training to obtain $M_3$. The full configuration is summarized in Table~\ref{tab:pruning_distillation_scaling_ladder}.

This construction offers two practical advantages for controlled experimentation. First, it substantially reduces training and evaluation cost, since $M_3$ is far less expensive to use than $M_0$. Second, because $M_3$ is derived from $M_0$ through a controlled sequence of pruning and distillation steps rather than trained independently from scratch, it retains important behavioral priors of the original large-scale model. This lineage makes observations on $M_3$ more representative of the large-model regime than those obtained from a randomly initialized model of comparable size. Taken together, the pruning--distillation pipeline serves as a cost-efficient scaling ladder for studying training objectives under controlled conditions.

\begin{table}[t]
\centering
\caption{Summary of the pruning--distillation scaling ladder. ``Parent Model'' denotes the model from which the current rung is pruned. ``Distillation Tokens'' denotes the number of tokens used for distillation-based continued pre-training after pruning.}
\label{tab:pruning_distillation_scaling_ladder}
\begin{tabular}{lcccc}
\toprule
\textbf{Parent Model} & \textbf{Resulting Ladder Rung} & \textbf{Total Parameters} & \textbf{Activated Parameters} & \textbf{Distillation Tokens} \\
\midrule
--    & $M_0$ & 671B & 37B  & --   \\
$M_0$ & $M_1$ & 128B & 19B  & 1T   \\
$M_1$ & $M_2$ & 59B  & 10B  & 200B \\
$M_2$ & $M_3$ & 24B  & 4.5B & 200B \\
\bottomrule
\end{tabular}
\end{table}

\subsubsection{Teacher and Student Models}
Unless otherwise specified, all subsequent experiments use $M_3$ as the student model. We use DeepSeek V3.1 Base as the teacher model, as it outperforms DeepSeek V3 Base according to our internal evaluations, making it the stronger teacher in our teacher--student training setup. The architecture configurations of both models are reported in Table~\ref{tab:teacher_student_architecture_config}.

\begin{table}[t]
\centering
\caption{Architecture configurations of the teacher and student models.}
\label{tab:teacher_student_architecture_config}
\begin{tabular}{lcc}
\toprule
\textbf{Configuration} & \textbf{Teacher: DeepSeek V3.1 Base} & \textbf{Student: \bm{$M_3$}} \\
\midrule
\texttt{num\_hidden\_layers} & 61 & 36 \\
\texttt{hidden\_size} & 7168 & 2880 \\
\texttt{intermediate\_size} & 18432 & 5184 \\
\texttt{moe\_intermediate\_size} & 2048 & 576 \\
\texttt{num\_attention\_heads} & 128 & 64 \\
\texttt{n\_routed\_experts} & 256 & 128 \\
\texttt{num\_experts\_per\_tok} & 8 & 8 \\
\bottomrule
\end{tabular}
\end{table}

\subsection{Training Data}
\label{subsec:training_data}
All training data used in both scaling-ladder construction and subsequent experiments are collected internally. The corpus combines open-source and internally synthetic data, processed through a unified internal data pipeline.

During scaling-ladder construction, we adopt a fixed data mixture across all stages, consisting of 50\% general-domain data, 30\% mathematics data, and 20\% code data. After the multi-stage pruning and distillation procedure summarized in Table~\ref{tab:pruning_distillation_scaling_ladder}, we evaluate $M_3$ following the protocol described in Section~\ref{subsec:evaluation_protocol} and report the corresponding results in Table~\ref{tab:motivating_observation_benchmark} and Table~\ref{tab:motivating_observation_pass_at_k}. The results indicate that $M_3$ achieves satisfactory performance as a pre-trained base model. Therefore in subsequent experiments, we increase the proportions of mathematics and code data relative to general-domain content. Concretely, the adjusted mixture consists of 33\% general-domain data, 40\% mathematics data, and 27\% code data.

\subsection{Evaluation Protocol}
\label{subsec:evaluation_protocol}

\subsubsection{Benchmark-Level Evaluation}
Following common practice, we evaluate models on a suite of 13 widely used benchmarks covering multiple capability dimensions assessed during pre-training, including factual knowledge, commonsense reasoning, code generation, and mathematical reasoning. Table~\ref{tab:pretrain_benchmarks} summarizes the benchmark suite, specifying the number of few-shot examples and the evaluation metric for each benchmark. To ensure robust performance tracking, we execute three evaluation runs for each benchmark and report the mean score across all trials.

\begin{table}[t]
\centering
\caption{Summary of the benchmark suite used for evaluating pre-trained base models. The table reports the benchmark name, the number of few-shot examples, and the evaluation metric.}
\label{tab:pretrain_benchmarks}
\begin{tabular}{lcc}
\toprule
\textbf{Benchmark} & \textbf{\# Few-shot} & \textbf{Metric} \\
\midrule
PIQA           & 0  & Acc    \\
HellaSwag      & 10 & Acc    \\
MMLU           & 5  & Acc    \\
TriviaQA       & 5  & EM     \\
RACE           & 5  & Acc    \\
DROP           & 3  & F1     \\
MMLU-Pro       & 5  & EM     \\
BBH            & 3  & EM     \\
C-Eval         & 5  & Acc    \\
CMMLU          & 5  & Acc    \\
MBPP           & 3  & Pass@1 \\
GSM8K          & 8  & EM     \\
MATH (Minerva) & 4  & EM     \\
\bottomrule
\end{tabular}
\end{table}

\subsubsection{Pass@$K$ Evaluation on Selected Benchmarks}
Single-sample benchmark metrics may not fully characterize a base model's capability, particularly for tasks where multiple generations can reveal whether the model places non-negligible probability mass on correct~\citep{goyal2026distilled}. To complement benchmark-level evaluation, we conduct zero-shot Pass@$K$ evaluation on selected benchmarks---MATH-500, AIME 2024, AIME 2025, and HumanEval---with $K \in \{1, 16, 64, 128\}$. The generation hyperparameters are fixed at top-$k=50$, top-$p=0.95$, and temperature $T=1.0$.

\section{Motivating Observation}

\subsection{Experimental Design}
\label{subsec:motivating_observation_experimental_design}

We begin with a controlled diagnostic experiment designed to isolate the effect of the training objective on model capabilities. The goal is not to maximize final model quality, but rather to characterize the differential effects of hard next-token supervision and soft teacher-distribution supervision under otherwise identical training conditions.

Following the experimental setup described in Section~\ref{sec:experimental_setup}, both variants are initialized from the same student checkpoint and trained on the same 100B-token corpus. All training hyperparameters are held fixed: a global batch size of 5760, a sequence length of 4096, and a constant learning rate of $5 \times 10^{-5}$. The sole difference between the two runs is the training objective. The \emph{LM} variant is optimized with the standard language modeling loss $L_{\mathrm{LM}}$. The \emph{KD} variant is optimized with the knowledge distillation loss $L_{\mathrm{KD}}$, using a teacher distribution constructed with top-$k$ truncation ($k = 256$) and temperature $\tau = 1$.

\subsection{Empirical Observations}

We evaluate both variants following the protocol described in Section~\ref{subsec:evaluation_protocol}. Table~\ref{tab:motivating_observation_benchmark} reports results on the standard benchmark suite. Table~\ref{tab:motivating_observation_pass_at_k} reports Pass@$K$ results.

\begin{table}[t]
\centering
\caption{Benchmark-level comparison between the LM and KD variants. $M_3$ denotes the warm-start model used to initialize both LM and KD training.}
\label{tab:motivating_observation_benchmark}
\begin{tabular}{lccc}
\toprule
\textbf{Benchmark} & \bm{$M_3$} & \textbf{LM} & \textbf{KD} \\
\midrule
PIQA           & 80.20 & 79.16 & \textbf{80.63} \\
HellaSwag      & 79.02 & 79.06 & \textbf{79.16} \\
MMLU           & 68.12 & 68.27 & \textbf{68.83} \\
TriviaQA       & 54.21 & 54.65 & \textbf{55.49} \\
RACE           & 43.73 & \textbf{46.89} & 45.74 \\
DROP           & 50.72 & 51.59 & \textbf{53.37} \\
MMLU-Pro       & 38.95 & \textbf{43.98} & 39.74 \\
BBH            & 66.96 & \textbf{69.90} & 69.33 \\
C-Eval         & 68.80 & \textbf{69.91} & 68.65 \\
CMMLU          & 72.03 & 72.41 & \textbf{72.72} \\
MBPP           & 50.60 & 53.00 & \textbf{54.80} \\
GSM8K          & 68.31 & 70.43 & \textbf{70.74} \\
MATH (Minerva) & 40.16 & \textbf{43.38} & 40.64 \\
\bottomrule
\end{tabular}
\end{table}

\begin{table*}[t]
\centering
\caption{Pass@$K$ comparison on mathematical reasoning and code-generation benchmarks. $M_3$ denotes the warm-start model used to initialize both LM and KD training.}
\label{tab:motivating_observation_pass_at_k}
\resizebox{\textwidth}{!}{
\begin{tabular}{lcccccccccccc}
\toprule
\multirow{2}{*}{\textbf{Benchmark}}
& \multicolumn{3}{c}{\textbf{Pass@1}}
& \multicolumn{3}{c}{\textbf{Pass@16}}
& \multicolumn{3}{c}{\textbf{Pass@64}}
& \multicolumn{3}{c}{\textbf{Pass@128}} \\
\cmidrule(lr){2-4}
\cmidrule(lr){5-7}
\cmidrule(lr){8-10}
\cmidrule(lr){11-13}
& \bm{$M_3$} & \textbf{LM} & \textbf{KD}
& \bm{$M_3$} & \textbf{LM} & \textbf{KD}
& \bm{$M_3$} & \textbf{LM} & \textbf{KD}
& \bm{$M_3$} & \textbf{LM} & \textbf{KD} \\
\midrule
MATH-500   & 3.00  & \textbf{13.02} & 3.15  & 29.97 & \textbf{60.75} & 30.89 & 56.49 & \textbf{78.80} & 57.41 & 68.00 & \textbf{84.40} & 68.40 \\
AIME 2024  & 0.03  & \textbf{0.57}  & 0.08  & 0.42  & \textbf{5.80}  & 1.25  & 1.67  & \textbf{14.17} & 5.00  & 3.33  & \textbf{23.33} & 10.00 \\
AIME 2025  & 0.03  & \textbf{0.23}  & 0.00  & 0.42  & \textbf{3.65}  & 0.00  & 1.67  & \textbf{13.35} & 0.00  & 3.33  & \textbf{23.33} & 0.00  \\
HumanEval  & 19.54 & \textbf{29.63} & 22.79 & 66.57 & \textbf{77.88} & 72.39 & 83.05 & \textbf{90.25} & 86.25 & 89.63 & \textbf{93.90} & 90.24 \\
\bottomrule
\end{tabular}
}
\end{table*}

The results show that neither objective is uniformly superior. Instead, LM and KD exhibit consistent, benchmark-dependent differences. On benchmarks requiring complex reasoning, mathematical problem solving, or exam-style knowledge assessment, LM consistently outperforms KD. The most pronounced benchmark-level gaps appear on MMLU-Pro, where LM exceeds KD by 4.24 points, and on MATH (Minerva), where the margin is 2.74 points. The Pass@$K$ evaluation further reinforces this pattern. On MATH-500, LM substantially outperforms KD across all sampling budgets, with gaps of 9.87 points at Pass@1 and 29.86 points at Pass@16. Consistent trends are observed on AIME 2024 and AIME 2025; on AIME 2025 in particular, the KD variant achieves a zero pass rate at every reported sampling budget, while LM performance improves steadily with additional samples. HumanEval exhibits the same directional pattern, though the gap narrows as $K$ increases. Taken together, these findings indicate that LM training yields a stronger capability profile than KD for high-difficulty reasoning and code-generation tasks.

In contrast, KD outperforms LM on a distinct set of benchmarks more closely associated with commonsense plausibility, factual retrieval, local reading comprehension, and structured program synthesis. Specifically, KD improves over LM by 1.47 points on PIQA, 0.84 points on TriviaQA, 1.78 points on DROP, and 1.80 points on MBPP. Although these margins are smaller than the largest LM-favored gaps, they are consistent in direction, indicating that KD is not uniformly inferior to LM.

Overall, these results reveal a systematic, non-uniform trade-off between the two training objectives. LM proves more favorable on high-difficulty reasoning and mathematical benchmarks, while KD proves more favorable on knowledge-intensive, commonsense, reading comprehension, and structured code-generation tasks.

\subsection{Research Questions}

The observations above motivate a more fine-grained analysis of the optimization signals induced by the two training objectives. Because both variants share the same initialization, training corpus, model architecture, and hyperparameter configuration, their systematically divergent benchmark behavior can be attributed solely to the choice of training objective. This gives rise to the first research question (RQ):

\paragraph{RQ1.}
How does top-$k$-truncated, temperature-scaled off-policy distillation reshape the token-level optimization landscape relative to standard LM training? Specifically, how do the truncation parameter $k$ and temperature $\tau$ govern the resulting token-level supervision signal and its downstream effects on benchmark performance?

The capability-dependent performance differences observed above indicate that LM and KD strengthen distinct capability dimensions. Committing to a single global training objective therefore amounts to fixing a particular trade-off among these dimensions---a choice that need not be optimal for all tasks of interest. This motivates the second research question:

\paragraph{RQ2.}
Can training objectives be routed adaptively across the training distribution---from coarse data subsets down to individual tokens---to differentially strengthen distinct capability dimensions, rather than committing to a single global trade-off?

\section{RQ1: How Truncation Size and Temperature Shape Token-Level Supervision in Off-Policy?}

This section addresses RQ1. We approach this question in four steps. We begin with a gradient-level comparison of $L_{\mathrm{KD}}$ and $L_{\mathrm{LM}}$ that establishes the structural difference at each token position. We then introduce a suite of diagnostic metrics to quantify key dimensions of the resulting supervision signal. Controlled sweeps over $k$ at fixed $\tau$ and over $\tau$ at fixed $k$ connect these token-level diagnostics to benchmark performance under each configuration. A concluding synthesis integrates the findings into a direct answer to the research question.

\subsection{Gradient-Level Comparison}
\label{subsec:gradient_level_comparison}

We begin by comparing the optimization signals induced by $L_{\mathrm{LM}}$ and $L_{\mathrm{KD}}$ at the level of student logits. Because both objectives are averaged over sequence positions, an analysis at a single position $r$ suffices; the sequence-level gradients differ only by a scalar factor of $1/n$. Throughout this subsection we omit the position index $r$ when there is no ambiguity.

Following the definitions in Section~\ref{sec:preliminaries}, $L_{\mathrm{LM}}$ uses the observed token $x$ as a one-hot target, whereas $L_{\mathrm{KD}}$ uses the top-$k$-truncated, temperature-scaled teacher distribution $p^{(T,k,\tau)}$. The partial derivatives with respect to the student logit $z_i^{(S)}$ are
\begin{equation}
\frac{\partial L_{\mathrm{LM}}}{\partial z_i^{(S)}}
=
p_i^{(S)} - \mathbf{1}[i=x],
\label{eq:lm_gradient}
\end{equation}
and
\begin{equation}
\frac{\partial L_{\mathrm{KD}}}{\partial z_i^{(S)}}
=
p_i^{(S)} - p_i^{(T,k,\tau)}.
\label{eq:kd_gradient}
\end{equation}
Subtracting~\eqref{eq:lm_gradient} from~\eqref{eq:kd_gradient}, the gradient gap between the two objectives is
\begin{equation}
\Delta g_i
\triangleq
\frac{\partial L_{\mathrm{KD}}}{\partial z_i^{(S)}}
-
\frac{\partial L_{\mathrm{LM}}}{\partial z_i^{(S)}}
=
\mathbf{1}[i=x] - p_i^{(T,k,\tau)}.
\label{eq:gradient_gap}
\end{equation}
The student distribution $p^{(S)}$ cancels in the difference, so $\Delta g_i$ depends only on the discrepancy between the one-hot data target and the truncated teacher target.

The structure of $\Delta g_i$ is governed entirely by whether the observed token falls within the teacher top-$k$ support $K^{(T)}$; we examine each case in turn.

\paragraph{Case~I: $x \in K^{(T)}$.}
When the observed token is covered by the teacher support, the gradient gap is
\begin{equation}
\Delta g_i
=
\begin{cases}
-p_i^{(T,k,\tau)},
& i \in K^{(T)},\ i \neq x, \\[4pt]
1 - p_x^{(T,k,\tau)},
& i = x, \\[4pt]
0,
& i \notin K^{(T)}.
\end{cases}
\label{eq:gradient_gap_x_in_support}
\end{equation}

For teacher-supported tokens $i \in K^{(T)} \setminus \{x\}$, the gap $\Delta g_i = -p_i^{(T,k,\tau)} < 0$ implies that the KD gradient is algebraically smaller than the LM gradient at these positions. Under gradient descent, this translates into a less suppressive update on the corresponding logits—or even a logit increase when $p_i^{(T,k,\tau)} > p_i^{(S)}$. These tokens are therefore treated as teacher-supported alternatives rather than uniformly suppressed examples.

For the observed token, $\Delta g_x = 1 - p_x^{(T,k,\tau)} > 0$ whenever $p_x^{(T,k,\tau)} < 1$, indicating that $L_{\mathrm{KD}}$ applies weaker upward pressure on $z_x^{(S)}$ than $L_{\mathrm{LM}}$. In other words, even when $x \in K^{(T)}$, $L_{\mathrm{KD}}$ dilutes token-level supervision of the observed token by redistributing part of the target mass to other teacher-supported alternatives.

Tokens outside the teacher support are assigned zero target mass by both objectives ($\Delta g_i = 0$); the two losses thus impose identical logit-level pressure on these tokens.

\paragraph{Case~II: $x \notin K^{(T)}$.} 
When the observed token falls outside the teacher support, $p_x^{(T,k,\tau)} = 0$, and the gradient gap becomes
\begin{equation}
\Delta g_i
=
\begin{cases}
-p_i^{(T,k,\tau)},
& i \in K^{(T)}, \\[4pt]
1,
& i = x, \\[4pt]
0,
& i \notin K^{(T)},\ i \neq x.
\end{cases}
\label{eq:gradient_gap_x_not_in_support}
\end{equation}

All teacher-supported tokens are now non-observed alternatives. As in Case~I, $L_{\mathrm{KD}}$ preserves or increases probability mass on these tokens relative to $L_{\mathrm{LM}}$.

The sharpest discrepancy concerns the observed token itself. Since $x \notin K^{(T)}$, $L_{\mathrm{KD}}$ assigns zero target probability to $x$, yielding
\begin{equation}
\frac{\partial L_{\mathrm{KD}}}{\partial z_x^{(S)}} = p_x^{(S)},
\end{equation}
whereas $L_{\mathrm{LM}}$ gives
\begin{equation}
\frac{\partial L_{\mathrm{LM}}}{\partial z_x^{(S)}} = p_x^{(S)} - 1.
\end{equation}
Under gradient descent, $L_{\mathrm{LM}}$ increases $z_x^{(S)}$, whereas $L_{\mathrm{KD}}$ decreases it. This sign reversal represents the most severe conflict between the two objectives: the observed token is a positive target under LM but receives no target mass under KD.

For tokens in neither $K^{(T)}$ nor $\{x\}$, the gradient gap is again zero.

This case analysis reveals the fundamental divergence between $L_{\mathrm{LM}}$ and $L_{\mathrm{KD}}$. The former concentrates all target mass on the observed token; the latter reallocates it to the teacher top-$k$ support. When $x \in K^{(T)}$, $L_{\mathrm{KD}}$ weakens the exclusive supervision of the observed token and treats teacher-supported alternatives as valid targets. When $x \notin K^{(T)}$, $L_{\mathrm{KD}}$ removes the positive token-level signal entirely and directly conflicts with $L_{\mathrm{LM}}$ on the observed token. These structural differences motivate the diagnostic metrics introduced in the following section.
\subsection{Diagnostic Metrics for Token-Level Supervision}
\label{subsec:diagnostic_metrics}

The gradient decomposition in Section~\ref{subsec:gradient_level_comparison} shows that the influence of $L_{\mathrm{KD}}$ on token-level supervision can be characterized along three complementary dimensions: whether the observed token falls within the teacher's top-$k$ support, how much target mass the teacher distribution assigns to the observed token, and how the probability mass is distributed among the teacher-supported set. We introduce a suite of diagnostic metrics to quantify each dimension.

\paragraph{Observed-token coverage.}
The first metric records whether $L_{\mathrm{KD}}$ assigns any positive target mass to the observed token:
\begin{equation}
\mathrm{Coverage}@k
=
\mathbf{1}
\!\left[
x_r \in K_r^{(T)}
\right].
\label{eq:coverage_at_k}
\end{equation}
This metric directly distinguishes the two cases analyzed in Eqs.~\eqref{eq:gradient_gap_x_in_support}--\eqref{eq:gradient_gap_x_not_in_support}. Because support membership depends solely on the top-$k$ token ranking, Coverage@$k$ is controlled by $k$ and is invariant to temperature scaling, provided that temperature does not alter the teacher's token ranking.

\paragraph{Observed-token rank.}
To obtain a more granular characterization of the observed token's position within the teacher distribution, we measure its rank under the teacher's next-token logits. Let $z_r^{(T)}(v)$ denote the teacher logit for token $v$ at position $r$. The observed-token rank is
\begin{equation}
\mathrm{OTRank}
=
1
+
\sum_{v \in \mathcal{V}}
\mathbf{1}
\!\left[
z_r^{(T)}(v) > z_r^{(T)}(x_r)
\right],
\label{eq:observed_token_rank}
\end{equation}
A lower value indicates that the observed token ranks closer to the top of the teacher distribution. Like coverage, OTRank is invariant to temperature scaling as long as the teacher's token ordering is preserved.

\paragraph{Observed-token mass.}
Coverage@$k$ indicates only whether the observed token is included in the teacher support; it does not quantify the strength of the resulting KD signal. We therefore define the observed-token mass as
\begin{equation}
\mathrm{OTMass}(k,\tau)
=
p_{r,x_r}^{(T,k,\tau)}.
\label{eq:observed_token_mass}
\end{equation}
Because $p_{r,x_r}^{(T,k,\tau)}=0$ whenever $x_r \notin K_r^{(T)}$, this metric jointly reflects support coverage and within-support mass allocation. A lower value indicates weaker observed-token reinforcement relative to $L_{\mathrm{LM}}$, which always assigns a target mass of one to $x_r$.

To disentangle the contribution of support coverage from within-support mass allocation, we also report the conditional observed-token mass:
\begin{equation}
\mathrm{CondOTMass}(k,\tau)
=
p_{r,x_r}^{(T,k,\tau)},
\qquad
\text{conditioned on } x_r \in K_r^{(T)}.
\label{eq:conditional_observed_token_mass}
\end{equation}
Unlike coverage and rank, both mass metrics are sensitive to $k$ and $\tau$, since both parameters affect the normalized teacher probabilities $p_{r,i}^{(T,k,\tau)}$.

\paragraph{Teacher target distribution statistics.}
Let
\[
p_{r,(1)}^{(T,k,\tau)}
\ge
p_{r,(2)}^{(T,k,\tau)}
\ge
\dots
\ge
p_{r,(k)}^{(T,k,\tau)}
\]
denote the order statistics of the truncated teacher distribution at position $r$. We summarize its concentration and softness with the following statistics.

The top-ranked teacher probability is
\begin{equation}
\mathrm{Top1Prob}(k,\tau)
=
p_{r,(1)}^{(T,k,\tau)}.
\label{eq:top1_prob}
\end{equation}
A high value indicates that the KD target is concentrated on the teacher's most probable token, approaching hard pseudo-labeling; a low value indicates that substantial mass is distributed over lower-ranked alternatives.

The entropy of the truncated teacher distribution is
\begin{equation}
\mathrm{Entropy}(k,\tau)
=
-
\sum_{i \in K_r^{(T)}}
p_{r,i}^{(T,k,\tau)}
\log p_{r,i}^{(T,k,\tau)}.
\label{eq:teacher_entropy}
\end{equation}
When comparing results across different values of $k$, we report the normalized entropy:
\begin{equation}
\mathrm{NormEntropy}(k,\tau)
=
\frac{
-
\sum_{i \in K_r^{(T)}}
p_{r,i}^{(T,k,\tau)}
\log p_{r,i}^{(T,k,\tau)}
}{
\log k
}.
\label{eq:normalized_teacher_entropy}
\end{equation}
These metrics characterize the extent to which the KD target spreads mass over teacher-supported alternatives. Under a fixed top-$k$ support, increasing temperature generally raises entropy by redistributing probability mass from the highest-ranked tokens toward lower-ranked alternatives.

Finally, we measure the pre-truncation top-$k$ probability mass, which quantifies how much of the original teacher distribution is retained by the top-$k$ support prior to renormalization. Let
\begin{equation}
\widetilde{p}_{r,i}^{(T,\tau)}
=
\frac{
\exp(z_{r,i}^{(T)} / \tau)
}{
\sum_{j \in \mathcal{V}}
\exp(z_{r,j}^{(T)} / \tau)
}
\label{eq:full_teacher_prob}
\end{equation}
denote the full-vocabulary teacher distribution before truncation. The pre-truncation top-$k$ mass is then
\begin{equation}
\mathrm{RawTopKMass}(k,\tau)
=
\sum_{i \in K_r^{(T)}}
\widetilde{p}_{r,i}^{(T,\tau)}.
\label{eq:raw_topk_mass}
\end{equation}
This quantity is distinct from the total mass of $p^{(T,k,\tau)}$ over $K_r^{(T)}$, which is identically one by construction; it instead reflects the fraction of teacher probability that truncation discards from the full vocabulary.

In the controlled experiments that follow, we use these metrics to characterize how varying $k$ at fixed $\tau$ and varying $\tau$ at fixed $k$ reshape the token-level supervision signal, and to connect the resulting shifts to downstream evaluation behavior.
\subsection{Effect of Top-$k$}
\label{subsec:effect_of_top_k}

\subsubsection{Experimental Setup}

We examine how the top-$k$ truncation parameter affects off-policy distillation. We vary the teacher support size over $k \in \{1, 4, 16, 64, 256\}$ while fixing the distillation temperature at $\tau=1$. All other training hyperparameters follow the configuration described in Section~\ref{subsec:motivating_observation_experimental_design}.

We evaluate each setting using the protocol described in Section~\ref{subsec:evaluation_protocol}. Table~\ref{tab:top_k_benchmark} reports results on the standard benchmark suite. Table~\ref{tab:topk_pass_at_k} further reports Pass@$K$ results, where $K$ denotes the number of sampled solutions at evaluation time and is distinct from the training-side truncation parameter $k$.

\begin{table}[t]
\centering
\caption{Benchmark results across teacher top-$k$ truncation settings ($\tau=1$).}
\label{tab:top_k_benchmark}
\begin{tabular}{lccccc}
\toprule
\textbf{Benchmark} & \textbf{Top-1} & \textbf{Top-4} & \textbf{Top-16} & \textbf{Top-64} & \textbf{Top-256} \\
\midrule
PIQA           & 78.67 & 80.03 & 79.98 & \textbf{80.90} & 80.63 \\
HellaSwag      & 74.28 & 78.77 & 78.83 & 78.97 & \textbf{79.16} \\
MMLU           & 68.62 & 67.89 & 68.03 & \textbf{69.19} & 68.83 \\
TriviaQA       & \textbf{56.06} & 55.18 & 55.10 & 55.66 & 55.49 \\
RACE           & 44.98 & 45.93 & 46.12 & \textbf{46.32} & 45.74 \\
DROP           & \textbf{54.76} & 54.05 & 52.33 & 53.40 & 53.37 \\
MMLU-Pro       & 40.11 & \textbf{42.53} & 42.42 & 41.27 & 39.74 \\
BBH            & 69.36 & 68.98 & \textbf{70.31} & 69.14 & 69.33 \\
C-Eval         & 67.98 & 69.02 & \textbf{69.32} & 69.17 & 68.65 \\
CMMLU          & 72.01 & 72.22 & 72.31 & \textbf{72.82} & 72.72 \\
MBPP           & \textbf{54.80} & 53.80 & 53.80 & 53.40 & \textbf{54.80} \\
GSM8K          & 69.98 & 70.51 & 69.52 & 69.29 & \textbf{70.74} \\
MATH (Minerva) & 40.66 & 42.08 & \textbf{42.40} & 40.08 & 40.64 \\
\bottomrule
\end{tabular}
\end{table}

\begin{table*}[t]
\centering
\caption{Pass@$K$ results across teacher top-$k$ truncation settings on mathematical reasoning and code-generation benchmarks.}
\label{tab:topk_pass_at_k}
\resizebox{\textwidth}{!}{
\begin{tabular}{lcccccccccccccccccccc}
\toprule
\multirow{2}{*}{\textbf{Benchmark}}
& \multicolumn{5}{c}{\textbf{Pass@1}}
& \multicolumn{5}{c}{\textbf{Pass@16}}
& \multicolumn{5}{c}{\textbf{Pass@64}}
& \multicolumn{5}{c}{\textbf{Pass@128}} \\
\cmidrule(lr){2-6}
\cmidrule(lr){7-11}
\cmidrule(lr){12-16}
\cmidrule(lr){17-21}
& \textbf{Top-1} & \textbf{Top-4} & \textbf{Top-16} & \textbf{Top-64} & \textbf{Top-256}
& \textbf{Top-1} & \textbf{Top-4} & \textbf{Top-16} & \textbf{Top-64} & \textbf{Top-256}
& \textbf{Top-1} & \textbf{Top-4} & \textbf{Top-16} & \textbf{Top-64} & \textbf{Top-256}
& \textbf{Top-1} & \textbf{Top-4} & \textbf{Top-16} & \textbf{Top-64} & \textbf{Top-256} \\
\midrule
MATH-500  & 5.10  & 8.79  & \textbf{9.24}  & 3.43  & 3.15  & 34.52 & 54.57 & \textbf{56.06} & 30.84 & 30.89 & 57.22 & 76.54 & \textbf{78.39} & 55.53 & 57.41 & 67.60 & 83.00 & \textbf{85.00} & 66.60 & 68.40 \\
AIME 2024 & 0.00  & 0.10  & \textbf{0.16}  & 0.03  & 0.08  & 0.00  & 1.62  & \textbf{2.45}  & 0.42  & 1.25  & 0.00  & 5.84  & \textbf{9.17}  & 1.67  & 5.00  & 0.00  & 10.00 & \textbf{16.67} & 3.33  & 10.00 \\
AIME 2025 & 0.00  & \textbf{0.18}  & 0.13  & 0.05  & 0.00  & 0.00  & \textbf{2.92}  & 2.08  & 0.78  & 0.00  & 0.00  & \textbf{11.67} & 8.33  & 2.51  & 0.00  & 0.00  & \textbf{23.33} & 16.67 & 3.33  & 0.00  \\
HumanEval & \textbf{37.22} & 27.92 & 27.74 & 22.38 & 22.79 & \textbf{77.85} & 76.22 & 75.78 & 70.83 & 72.39 & 86.43 & 87.61 & \textbf{87.72} & 86.39 & 86.25 & 89.63 & 90.85 & 90.85 & \textbf{91.46} & 90.24 \\
\bottomrule
\end{tabular}
}
\end{table*}

\subsubsection{Analysis}

Scores on many benchmarks vary only marginally across top-$k$ settings. To focus the analysis on benchmarks with non-trivial sensitivity, we compute the relative score range
\[
    \frac{|\max_k s_k - \min_k s_k|}{\mathrm{avg}_k(s_k)},
\]
where $s_k$ denotes the score under top-$k$ distillation, and use this ratio to identify benchmarks with meaningful top-$k$ sensitivity, setting aside those whose variation is likely dominated by measurement noise.

\paragraph{Observed benchmark patterns.}
The standard benchmark suite exhibits heterogeneous responses to teacher support size. HellaSwag and PIQA clearly benefit from expanding the support beyond Top-1, and RACE shows a milder trend in the same direction. In contrast, MMLU-Pro and MATH (Minerva) favor intermediate support sizes---typically Top-4 or Top-16---and degrade as the support grows to Top-256. DROP exhibits yet another pattern: its best score occurs under Top-1, suggesting that this benchmark may benefit more from a sharper target distribution than from additional teacher-supported alternatives. Overall, these results indicate that the effect of top-$k$ truncation is non-monotonic and task-dependent; no single support size dominates across the full benchmark suite.

Pass@$K$ evaluations show stronger top-$k$ sensitivity than the standard benchmark suite, particularly on mathematical reasoning tasks. On MATH-500 and AIME, Top-4 and Top-16 outperform both the narrow Top-1 setting and the broader Top-64 and Top-256 settings. HumanEval exhibits a different pattern: it favors Top-1 at Pass@1, but the gap across top-$k$ settings narrows substantially as the sampling budget increases. This suggests that top-$k$ has a more pronounced influence on difficult mathematical reasoning tasks that require large sampling budgets, whereas code generation becomes less sensitive once a sufficient number of candidate solutions are drawn.

\begin{figure}[t]
    \centering
    \includegraphics[width=\linewidth]{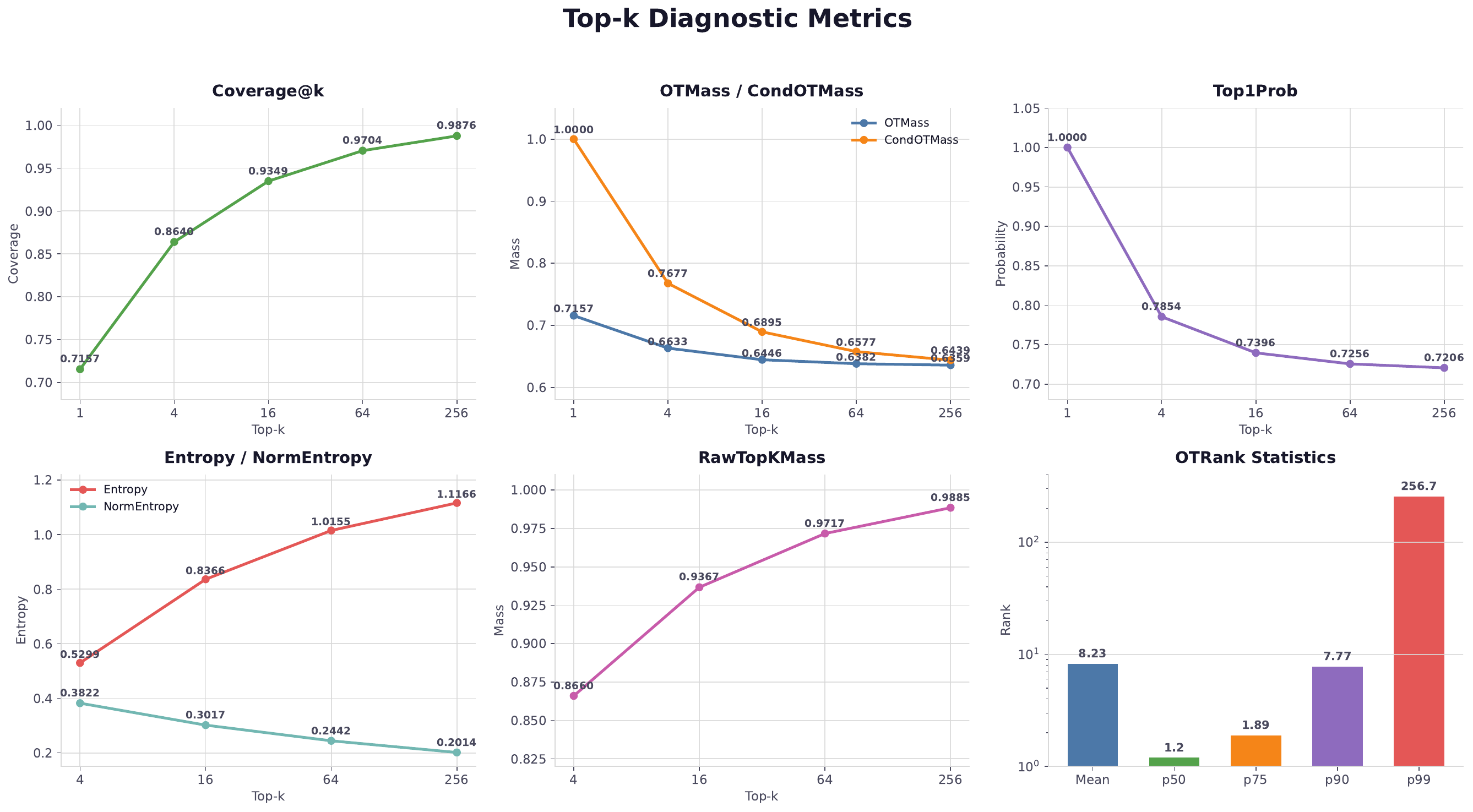}
    \caption{
        Diagnostic statistics for top-$k$ truncated teacher distributions at $\tau=1$.
    }
    \label{fig:topk_diagnostic_metrics}
\end{figure}

\paragraph{Diagnostic statistics.}
To analyze these patterns, we randomly sample 1B tokens from the training corpus and report the position-averaged diagnostic statistics defined in Section~\ref{subsec:diagnostic_metrics}. Figure~\ref{fig:topk_diagnostic_metrics} summarizes the resulting token-level measurements.

The diagnostics reveal a systematic trade-off induced by increasing the teacher support size. Observed-token coverage increases monotonically with $k$: Coverage@$k$ rises from $0.7157$ at Top-1 to $0.8640$, $0.9349$, $0.9704$, and $0.9876$ at Top-4, Top-16, Top-64, and Top-256, respectively. Larger support sizes therefore reduce the frequency with which the observed data token falls outside the teacher support.

This coverage gain is accompanied by weaker observed-token reinforcement. OTMass decreases from $0.7157$ at Top-1 to $0.6359$ at Top-256, while CondOTMass falls from $1.0000$ to $0.6439$. Top1Prob declines from $1.0000$ to $0.7206$, and the entropy of the truncated distribution rises from $0.5299$ at Top-4 to $1.1166$ at Top-256. Because raw entropy partly depends on support size, we interpret it jointly with Top1Prob, OTMass, CondOTMass, and normalized entropy. Taken together, these statistics indicate that larger support sizes produce a less concentrated KD target: more observed tokens are covered, but each observed token receives less probability mass, and the remaining mass is spread more diffusely over teacher-supported alternatives.

Observed-token rank statistics clarify why the marginal coverage benefit of increasing $k$ diminishes rapidly. The median observed-token rank is close to $1$, the 75th percentile falls below $2$, and the 90th percentile is only $7.77$, while the 99th percentile reaches $256.70$. Consequently, the vast majority of observed tokens already lie within a small or intermediate support such as Top-4 or Top-16. Increasing $k$ beyond this range primarily extends the support to long-tail positions and introduces additional low-probability teacher alternatives without meaningfully improving coverage for most tokens.

\paragraph{Core mechanism.}
The diagnostics suggest that the top-$k$ parameter governs a fundamental coverage--sharpness trade-off. Increasing $k$ raises observed-token coverage, reducing support mismatch, but simultaneously lowers observed-token mass and raises entropy, producing a progressively softer target distribution over teacher-supported alternatives. The coverage gain can be beneficial when Top-1 is too restrictive and frequently excludes observed tokens from the teacher support. The resulting target softening, however, can be detrimental when the target capability requires precise token-level reinforcement, as the KD objective becomes increasingly diffuse over teacher-preferred alternatives.

Because the diagnostic statistics are computed on a random sample from the training corpus rather than on benchmark-specific distributions, they characterize the global optimization signal induced by top-$k$ truncation. The benchmark-level interpretations that follow should therefore be understood as hypotheses consistent with the diagnostics rather than as direct causal attributions.

\paragraph{Analysis of standard benchmarks.}
The aggregate score on the standard benchmark suite is relatively insensitive to top-$k$, which is consistent with the diagnostic trade-off. Expanding the support beyond Top-1 improves observed-token coverage but also reduces CondOTMass and softens the target distribution. Across a diverse benchmark suite, these two effects tend to partially offset each other: some tasks may benefit from reduced support mismatch, while others may be adversely affected by weaker observed-token reinforcement. As a result, no single top-$k$ value dominates uniformly across the standard benchmarks.

HellaSwag and PIQA---with RACE as a milder case---are representative benchmarks that benefit from larger support sizes. Under Top-1, Coverage@$k$ is $0.7157$, meaning a substantial fraction of observed tokens receive zero KD target mass when they fall outside the teacher's top prediction. Expanding the support to Top-4 or Top-16 raises coverage sharply to $0.8640$ and $0.9349$, respectively. For commonsense plausibility and reading-style multiple-choice tasks, the additional teacher-supported alternatives may preserve semantically relevant uncertainty rather than acting as noise. In these cases, the coverage gain and richer soft supervision appear to outweigh the dilution of observed-token mass.

MMLU-Pro and MATH (Minerva) exhibit a different pattern. Both benefit from moving beyond Top-1 but degrade as the support becomes too broad, consistent with an intermediate optimum in the coverage--sharpness trade-off. Top-4 and Top-16 substantially reduce support mismatch while maintaining a relatively concentrated target distribution. Expanding to Top-64 or Top-256 yields only modest additional coverage while further reducing CondOTMass and increasing entropy. For knowledge-intensive and mathematical reasoning tasks, this additional softening may weaken the precise token-level optimization signal required for stable reasoning chains and exact answer formation.

DROP provides a useful counterexample. Its preference for Top-1 suggests that certain reading-comprehension and numerical QA settings may benefit more from a sharp pseudo-label-like target than from a broader teacher distribution. A plausible explanation is that DROP-style tasks---which often involve localized evidence selection, span matching, or numerical manipulation---are sensitive to the dilution of exact-token reinforcement by additional soft supervision. This observation reinforces the conclusion that larger teacher support is not universally beneficial.

\paragraph{Analysis of Pass@$K$ results.}
The Pass@$K$ results, particularly on mathematical reasoning benchmarks, are more sensitive to top-$k$ than the standard benchmark suite. A plausible explanation is that long-chain reasoning amplifies both token-level support mismatch and the effect of target softening. Under Top-1, many intermediate reasoning tokens may fall outside the teacher support, and local supervision gaps can accumulate over long solution trajectories. Expanding the support to Top-4 or Top-16 substantially improves observed-token coverage, thereby reducing this mismatch.

However, excessively large supports are also suboptimal. From Top-16 to Top-256, Coverage@$k$ increases only from $0.9349$ to $0.9876$, while CondOTMass decreases from $0.6895$ to $0.6439$ and entropy continues to rise. The additional support thus primarily introduces low-probability alternatives rather than meaningfully improving coverage for most observed tokens. For MATH-500 and AIME, where successful generation requires coherent and precise reasoning trajectories, a progressively softer target may reduce the probability of producing valid solution paths. This interpretation is consistent with the finding that Top-4 and Top-16 outperform both Top-1 and Top-64/Top-256 on these mathematical Pass@$K$ evaluations.

HumanEval behaves differently from the mathematical benchmarks. Its Pass@1 score favors Top-1, suggesting that sharp supervision may aid single-sample code generation, where syntactic correctness, API usage, and formatting precision are critical. As the sampling budget increases, the performance gap across top-$k$ settings narrows. Under larger sampling budgets, success depends less on the single highest-probability generation and more on whether the model can produce at least one correct candidate, which may partially offset the advantage of the sharper Top-1 target.

\paragraph{Why Top-64 and Top-256 perform similarly.}
The small performance difference between Top-64 and Top-256 follows directly from the diagnostics. By Top-64, Coverage@$k$ has already reached $0.9704$ and RawTopKMass is $0.9717$. Increasing the support to Top-256 raises these figures only to $0.9876$ and $0.9885$, respectively, with correspondingly modest changes in OTMass, CondOTMass, and Top1Prob. Top-256 therefore adds predominantly low-probability tail tokens, yielding only a marginal shift in the effective optimization signal for most positions. These considerations help explain why Top-64 and Top-256 often yield similar performance on the standard benchmark suite, while leaving room for larger divergences in settings that are particularly sensitive to target softening, such as low-budget mathematical Pass@$K$ evaluations.
\subsection{Effect of Temperature}
\label{subsec:effect_of_temperature}

\subsubsection{Experimental Setup}

We next study how distillation temperature affects off-policy distillation. Based on the findings in Section~\ref{subsec:effect_of_top_k}, we fix the teacher support size at $k=256$ and vary the temperature over $\tau \in \{0.25, 0.5, 1.0\}$. This broad-support regime ensures that the teacher support covers most observed tokens and retains most of the raw teacher probability mass, so the temperature sweep isolates within-support probability allocation rather than support coverage effects.

We also include the limiting case $\tau \to 0$ as a hard pseudo-labeling baseline. In this limit, the top-$256$ teacher distribution collapses to a one-hot target on the teacher's top-1 token and is therefore equivalent to top-$1$ distillation in terms of the resulting target distribution, absent ties in the teacher logits. All other training hyperparameters follow the configuration described in Section~\ref{subsec:motivating_observation_experimental_design}.

We evaluate each temperature setting using the protocol described in Section~\ref{subsec:evaluation_protocol}. Table~\ref{tab:temperature_benchmark} reports results on the standard benchmark suite. Table~\ref{tab:temperature_pass_at_k} further reports Pass@$K$ results.

\begin{table}[t]
\centering
\caption{Benchmark results under different distillation temperatures. The limiting case $\tau \to 0$ collapses the teacher target to a one-hot distribution on the teacher's top-1 token, equivalent to top-$1$ distillation in terms of the resulting target distribution.}
\label{tab:temperature_benchmark}
\begin{tabular}{lcccc}
\toprule
\textbf{Benchmark} 
& \bm{$\tau \to 0$} 
& \bm{$\tau=0.25$} 
& \bm{$\tau=0.5$} 
& \bm{$\tau=1.0$} \\
\midrule
PIQA           & 78.67 & 79.82 & 80.14 & \textbf{80.63} \\
HellaSwag      & 74.28 & 75.08 & 76.87 & \textbf{79.16} \\
MMLU           & 68.62 & 68.73 & \textbf{69.25} & 68.83 \\
TriviaQA       & 56.06 & \textbf{56.28} & 56.15 & 55.49 \\
RACE           & 44.98 & 45.74 & \textbf{45.84} & 45.74 \\
DROP           & \textbf{54.76} & 54.43 & 54.61 & 53.37 \\
MMLU-Pro       & 40.11 & \textbf{40.87} & 40.31 & 39.74 \\
BBH            & 69.36 & 68.99 & \textbf{69.57} & 69.33 \\
C-Eval         & 67.98 & 67.90 & \textbf{68.95} & 68.65 \\
CMMLU          & 72.01 & 72.40 & \textbf{72.77} & 72.72 \\
MBPP           & \textbf{54.80} & 52.40 & 52.00 & \textbf{54.80} \\
GSM8K          & 69.98 & 69.90 & 69.90 & \textbf{70.74} \\
MATH (Minerva) & 40.66 & \textbf{41.72} & 41.70 & 40.64 \\
\bottomrule
\end{tabular}
\end{table}

\begin{table*}[t]
\centering
\caption{Pass@$K$ results under different distillation temperatures on mathematical reasoning and code-generation benchmarks.}
\label{tab:temperature_pass_at_k}
\resizebox{\textwidth}{!}{
\begin{tabular}{lcccccccccccccccc}
\toprule
\multirow{2}{*}{\textbf{Benchmark}}
& \multicolumn{4}{c}{\textbf{Pass@1}}
& \multicolumn{4}{c}{\textbf{Pass@16}}
& \multicolumn{4}{c}{\textbf{Pass@64}}
& \multicolumn{4}{c}{\textbf{Pass@128}} \\
\cmidrule(lr){2-5}
\cmidrule(lr){6-9}
\cmidrule(lr){10-13}
\cmidrule(lr){14-17}
& \bm{$\tau \to 0$} & \bm{$\tau=0.25$} & \bm{$\tau=0.5$} & \bm{$\tau=1.0$}
& \bm{$\tau \to 0$} & \bm{$\tau=0.25$} & \bm{$\tau=0.5$} & \bm{$\tau=1.0$}
& \bm{$\tau \to 0$} & \bm{$\tau=0.25$} & \bm{$\tau=0.5$} & \bm{$\tau=1.0$}
& \bm{$\tau \to 0$} & \bm{$\tau=0.25$} & \bm{$\tau=0.5$} & \bm{$\tau=1.0$} \\
\midrule
MATH-500  & 4.88  & \textbf{5.44}  & 5.24  & 3.15  & 32.47 & 36.69 & \textbf{37.94} & 30.89 & 52.54 & 58.03 & \textbf{60.90} & 57.41 & 61.20 & 66.60 & \textbf{69.20} & 68.40 \\
AIME 2024 & 0.00  & 0.03  & \textbf{0.08}  & \textbf{0.08}  & 0.00  & 0.42  & \textbf{1.25}  & \textbf{1.25}  & 0.00  & 1.67  & \textbf{5.00}  & \textbf{5.00}  & 0.00  & 3.33  & \textbf{10.00} & \textbf{10.00} \\
AIME 2025 & 0.00  & 0.03  & \textbf{0.10}  & 0.00  & 0.00  & 0.42  & \textbf{1.67}  & 0.00  & 0.00  & 1.67  & \textbf{6.67}  & 0.00  & 0.00  & 3.33  & \textbf{13.33} & 0.00  \\
HumanEval & \textbf{37.22} & 35.03 & 32.18 & 22.79 & 77.85 & \textbf{78.18} & 76.23 & 72.39 & 86.43 & \textbf{87.44} & 86.68 & 86.25 & 89.63 & \textbf{90.24} & 89.63 & \textbf{90.24} \\
\bottomrule
\end{tabular}
}
\end{table*}

\subsubsection{Analysis}

\paragraph{Observed benchmark patterns.}
The standard benchmark suite exhibits heterogeneous responses to temperature. HellaSwag shows the clearest preference for higher temperature, with scores increasing from $74.28$ under $\tau \to 0$ to $79.16$ under $\tau=1.0$. PIQA follows the same direction, rising from $78.67$ to $80.63$. DROP exhibits the opposite pattern, achieving its best score under $\tau \to 0$ and its lowest under $\tau=1.0$. MMLU-Pro and MATH (Minerva) favor low-to-moderate temperatures: MMLU-Pro peaks at $\tau=0.25$, while MATH (Minerva) achieves its highest scores at $\tau=0.25$ and $\tau=0.5$. MBPP exhibits a less interpretable pattern---both $\tau \to 0$ and $\tau=1.0$ tie for the best score while intermediate temperatures underperform---and we therefore set it aside in the analysis below.

Pass@$K$ evaluations show stronger temperature sensitivity than the standard benchmark suite, particularly on mathematical reasoning tasks. For MATH-500, Pass@16, Pass@64, and Pass@128 all favor the intermediate setting $\tau=0.5$. AIME exhibits a similar pattern; on AIME 2025, Pass@128 peaks at $\tau=0.5$ while both $\tau \to 0$ and $\tau=1.0$ perform substantially worse. HumanEval follows a different pattern: Pass@1 strongly favors $\tau \to 0$, decreasing from $37.22$ to $22.79$ as $\tau$ increases to $1.0$, while Pass@64 and Pass@128 are nearly insensitive to temperature.

\begin{figure}[t]
    \centering
    \includegraphics[width=\linewidth]{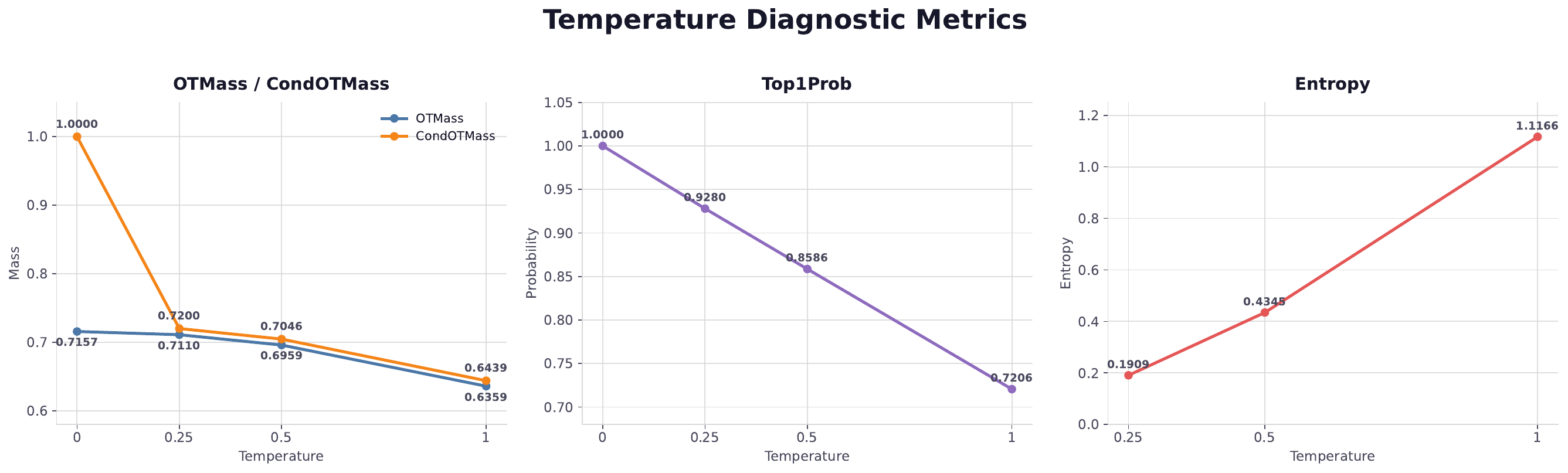}
    \caption{
        Diagnostic statistics for top-$256$ truncated teacher distributions under different distillation temperatures.
    }
    \label{fig:temperature_diagnostic_metrics}
\end{figure}

\paragraph{Diagnostic statistics.}
Figure~\ref{fig:temperature_diagnostic_metrics} reports the diagnostic statistics defined in Section~\ref{subsec:diagnostic_metrics}, computed on a random 1B-token sample from the training corpus.

Because temperature scaling preserves the teacher's token ranking, the top-$256$ support, Coverage@$k$, and observed-token rank are all invariant across this sweep. Temperature therefore affects only within-support probability allocation. As $\tau$ increases from $0$ to $1.0$, OTMass decreases from $0.7157$ to $0.6359$, and CondOTMass falls more sharply from $1.0000$ to $0.6439$. Top1Prob similarly declines from $1.0000$ to $0.7206$, while the entropy of the truncated teacher distribution rises from $0.1909$ at $\tau=0.25$ to $1.1166$ at $\tau=1.0$. These trends confirm that higher temperature redistributes probability mass from the top-ranked token toward lower-ranked teacher-supported alternatives.

\paragraph{Core mechanism.}
Under fixed top-$k$ truncation, temperature controls within-support probability allocation. At very low $\tau$, the teacher target approaches a one-hot pseudo-label on the teacher's top-1 token, providing strong direct reinforcement of the highest-ranked prediction while suppressing teacher-supported alternatives. As $\tau$ increases, the target distribution softens: mass flows from the top-ranked token toward lower-ranked alternatives within the support, reducing the average probability assigned to the observed token.

This induces a trade-off between two complementary training signals: direct observed-token reinforcement and teacher-supported alternative supervision. Lower temperature emphasizes sharp token-level supervision, resembling pseudo-label training. Higher temperature exposes the student to a broader set of teacher-preferred alternatives but simultaneously weakens direct reinforcement of the observed token. As in Section~\ref{subsec:effect_of_top_k}, the diagnostic statistics characterize the global optimization signal; benchmark-level interpretations below should be understood as hypotheses consistent with the diagnostics rather than as direct causal attributions.

\paragraph{Analysis of standard benchmarks.}
HellaSwag's improvement under higher temperature is consistent with the interpretation that commonsense plausibility tasks benefit from richer soft supervision. As a semantic discrimination benchmark, HellaSwag may contain positions where non-top-1 teacher-supported tokens encode locally plausible continuations or distributional preferences rather than noise. As temperature increases and these alternatives receive greater probability mass, the resulting training signal provides richer guidance than a near-one-hot target. PIQA exhibits a similar, albeit weaker, trend.

DROP exhibits the opposite behavior, performing best under the sharpest target and degrading at $\tau=1.0$. Reading-comprehension and numerical QA tasks often require precise evidence tokens, span-level grounding, or numerical specificity, concentrating relevant supervision on a single correct token or phrase. High-temperature teacher targets can dilute this signal by distributing more mass to within-support alternatives, consistent with the observed decreases in OTMass and CondOTMass.

MMLU-Pro and MATH (Minerva) favor intermediate temperatures. We hypothesize that these benchmarks benefit from a balance between target sharpness and limited alternative supervision: very low temperature over-concentrates the target on the teacher's top-1 prediction and suppresses useful alternative reasoning tokens, while high temperature weakens precise reasoning- and answer-token supervision. Intermediate temperatures ($\tau=0.25$--$0.5$) preserve a relatively concentrated target while allowing some non-top-1 teacher preferences to contribute to training.

\paragraph{Analysis of Pass@$K$ results.}
The Pass@$K$ results reveal stronger temperature sensitivity for mathematical generation than the standard benchmark suite. For MATH-500 and AIME, intermediate temperatures---particularly $\tau=0.5$---are consistently favorable under larger sampling budgets. A plausible interpretation is that successful mathematical generation benefits from learning a distribution over multiple reasoning trajectories: a near-one-hot target under $\tau \to 0$ may provide strong local supervision but underrepresent alternative reasoning paths that become useful under repeated sampling, while $\tau=1.0$ may be too diffuse to maintain precise reasoning-step supervision. The intermediate setting $\tau=0.5$ appears to balance these competing requirements.

HumanEval follows a different pattern. Its Pass@1 score clearly favors $\tau \to 0$, consistent with code generation's sensitivity to local syntax, API conventions, indentation, and variable naming, where a high-temperature teacher target may assign non-negligible mass to locally plausible but functionally incorrect alternatives. This advantage diminishes substantially at higher sampling budgets: Pass@64 and Pass@128 on HumanEval are nearly insensitive to temperature. This distinguishes HumanEval from the mathematical benchmarks, for which intermediate temperature remains advantageous even under large sampling budgets.
\subsection{Summary of Findings}
\label{subsec:answer_to_rq1}

The preceding analysis establishes that top-$k$-truncated, temperature-scaled off-policy distillation reshapes the token-level optimization landscape by replacing the one-hot supervision of standard LM training with a distributional teacher target defined over a restricted support. At the gradient level, the discrepancy between $L_{\mathrm{KD}}$ and $L_{\mathrm{LM}}$ depends solely on the difference between the observed data token and the truncated teacher distribution. When the observed token lies within the top-$k$ teacher support, $L_{\mathrm{KD}}$ retains positive target mass for that token, but the resulting upward gradient pressure is weaker than under $L_{\mathrm{LM}}$ because part of the probability mass is reallocated to teacher-supported alternatives. These alternatives are no longer treated as uniformly negative examples; they instead receive positive or less negative supervision. When the observed token falls outside the teacher support, the conflict is more severe: $L_{\mathrm{LM}}$ increases the corresponding logit, whereas $L_{\mathrm{KD}}$ assigns it zero target mass and suppresses it under gradient descent. Relative to standard LM training, $L_{\mathrm{KD}}$ thus trades exclusive observed-token reinforcement for teacher-supported alternative supervision, with the nature and severity of this trade-off determined by the support structure and its probability allocation.

The truncation size $k$ and temperature $\tau$ regulate this trade-off through different mechanisms. The top-$k$ parameter primarily governs a coverage--sharpness trade-off. Increasing $k$ raises the probability that the observed token falls within the teacher support, reducing support mismatch and thereby avoiding positions where $L_{\mathrm{KD}}$ assigns zero target mass to the data token. Coverage@$k$ and the observed-token rank distribution directly quantify this effect. However, expanding the support simultaneously dilutes the teacher target: OTMass and CondOTMass decrease, Top1Prob declines, and entropy rises, indicating that probability mass is distributed over a progressively broader set of teacher-supported alternatives. Temperature, by contrast, does not alter support membership or observed-token rank at fixed $k$, since rescaling logits preserves the teacher's token ordering. Instead, $\tau$ controls within-support probability allocation: lower temperature concentrates mass toward the top-ranked token, making the teacher target approach a hard pseudo-label, while higher temperature redistributes mass toward lower-ranked alternatives within the same support. Raising $\tau$ therefore weakens direct observed-token reinforcement at positions where the observed token is among the teacher's highest-ranked predictions, while strengthening the supervision carried by non-top-1 teacher-supported alternatives.

These two hyperparameters modulate two complementary factors: \emph{direct observed-token reinforcement} and \emph{teacher-supported alternative supervision}. Direct observed-token reinforcement is strongest when the teacher target is concentrated and the observed token receives high probability mass---as under Top-1 distillation, or under very low temperature when the observed token coincides with the teacher's top-ranked prediction. Teacher-supported alternative supervision is strongest when the support is broad or the temperature is high, exposing the student to a wider set of teacher-preferred continuations. Neither factor is universally beneficial. Strong observed-token reinforcement preserves precise data-token learning but may discard useful teacher uncertainty and suppress valid alternative continuations. Conversely, rich alternative supervision can expose the student to semantically or procedurally relevant teacher preferences, but may weaken exact-token supervision in tasks that require precise grounding, stable reasoning chains, or exact answer formation.

The benchmark results are consistent with this account and reveal considerable task heterogeneity. Commonsense plausibility tasks such as HellaSwag and PIQA benefit from broader or softer teacher supervision: scores improve when expanding beyond Top-1 and tend to increase with temperature. This pattern suggests that teacher-supported alternatives encode meaningful semantic uncertainty for these tasks rather than acting as noise. In contrast, DROP favors sharper targets, performing best under Top-1 or very low temperature, consistent with its requirement for localized evidence selection, span grounding, and numerical specificity---settings in which spreading probability mass over alternatives dilutes the exact-token signal. MMLU-Pro and MATH (Minerva) favor intermediate configurations---Top-4 or Top-16 at low-to-moderate temperatures---indicating that these tasks benefit from reduced support mismatch while still maintaining a sufficiently concentrated target.

The Pass@$K$ results reinforce and extend this picture. Mathematical generation benchmarks such as MATH-500 and AIME exhibit strong sensitivity to the coverage--sharpness balance: intermediate support sizes (Top-4 or Top-16) and intermediate temperatures (e.g., $\tau=0.5$) consistently outperform both highly concentrated and highly diffuse targets. This suggests that mathematical reasoning benefits from limited alternative supervision---plausibly because multiple valid reasoning trajectories are useful under large sampling budgets---while still requiring sufficient target concentration to sustain coherent derivations and precise answer tokens. HumanEval shows a different pattern: Pass@1 clearly favors sharper supervision, consistent with the sensitivity of code generation to local syntactic correctness, API usage, and formatting conventions, whereas sensitivity to $k$ and $\tau$ diminishes substantially at larger sampling budgets. Taken together, these findings indicate that top-$k$ and temperature do not act as simple quality multipliers for knowledge distillation; rather, they position the token-level optimization signal along a spectrum between direct observed-token reinforcement and teacher-supported alternative supervision, and different tasks favor different positions along this spectrum for optimal knowledge transfer.

\section{RQ2: Can Training Objectives Be Routed Adaptively Across the Training Distribution?}

We approach RQ2 through a sequence of progressively finer-grained analyses. The complementary performance profiles of $L_{\mathrm{LM}}$ and $L_{\mathrm{KD}}$ across benchmarks motivate a natural first step: assigning different training objectives to different data subsets according to coarse domain labels. To understand why domain-level routing may serve as an effective proxy for objective routing, we analyze the diagnostic metrics Coverage@$k$, CondOTMass, and Entropy stratified by domain, linking the coarse domain partition to the underlying token-level supervision statistics. Building on these findings, we evaluate finer-grained token-level routing policies that use OTMass and Entropy directly as per-token routing criteria. The section concludes with a direct answer to RQ2 that integrates the subset-level and token-level results, with particular attention to how routing-signal quality determines the practical effectiveness of adaptive objective routing.

\subsection{Domain-Based Subset-Level Objective Routing}
\label{subsec:domain_routing}

\begin{table}[t]
\centering
\caption{Benchmark-level comparison under different training-objective routing strategies.}
\label{tab:domain_routing_benchmark}
\begin{tabular}{lccc}
\toprule
\textbf{Benchmark} & \textbf{LM} & \textbf{KD} & \textbf{Domain} \\
\midrule
PIQA           & 79.16 & \textbf{80.63} & 80.36 \\
HellaSwag      & 79.06 & 79.16 & \textbf{79.30} \\
MMLU           & 68.27 & \textbf{68.83} & 68.12 \\
TriviaQA       & 54.65 & \textbf{55.49} & 55.33 \\
RACE           & \textbf{46.89} & 45.74 & 44.98 \\
DROP           & 51.59 & \textbf{53.37} & \textbf{53.37} \\
MMLU-Pro       & \textbf{43.98} & 39.74 & 43.72 \\
BBH            & 69.90 & 69.33 & \textbf{70.74} \\
C-Eval         & \textbf{69.91} & 68.65 & 68.80 \\
CMMLU          & 72.41 & \textbf{72.72} & 72.67 \\
MBPP           & 53.00 & 54.80 & \textbf{58.40} \\
GSM8K          & 70.43 & \textbf{70.74} & 70.58 \\
MATH (Minerva) & 43.38 & 40.64 & \textbf{43.58} \\
\bottomrule
\end{tabular}
\end{table}

\begin{table}[t]
\centering
\caption{Pass@$K$ comparison under different training-objective routing strategies.}
\label{tab:domain_routing_pass_at_k}
\resizebox{\textwidth}{!}{
\begin{tabular}{lccc ccc ccc ccc}
\toprule
\multirow{2}{*}{\textbf{Benchmark}}
& \multicolumn{3}{c}{\textbf{Pass@1}}
& \multicolumn{3}{c}{\textbf{Pass@16}}
& \multicolumn{3}{c}{\textbf{Pass@64}}
& \multicolumn{3}{c}{\textbf{Pass@128}} \\
\cmidrule(lr){2-4}
\cmidrule(lr){5-7}
\cmidrule(lr){8-10}
\cmidrule(lr){11-13}
& \textbf{LM} & \textbf{KD} & \textbf{Domain}
& \textbf{LM} & \textbf{KD} & \textbf{Domain}
& \textbf{LM} & \textbf{KD} & \textbf{Domain}
& \textbf{LM} & \textbf{KD} & \textbf{Domain} \\
\midrule
AIME24
& \textbf{0.57} & 0.08 & 0.52
& 5.80 & 1.25 & \textbf{6.62}
& 14.17 & 5.00 & \textbf{19.17}
& 23.33 & 10.00 & \textbf{33.33} \\
AIME25
& 0.23 & 0.00 & \textbf{0.39}
& 3.65 & 0.00 & \textbf{5.42}
& 13.35 & 0.00 & \textbf{15.39}
& \textbf{23.33} & 0.00 & \textbf{23.33} \\
MATH500
& \textbf{13.02} & 3.07 & 10.92
& \textbf{60.75} & 28.53 & 56.66
& \textbf{78.80} & 51.90 & 76.38
& \textbf{84.40} & 63.00 & 83.20 \\
HumanEval
& \textbf{29.63} & 22.79 & 28.63
& \textbf{77.88} & 72.39 & 76.88
& \textbf{90.25} & 86.25 & 89.18
& \textbf{93.90} & 90.24 & 92.68 \\
\bottomrule
\end{tabular}
}
\end{table}

The findings from RQ1 establish that $L_{\mathrm{LM}}$ and $L_{\mathrm{KD}}$ exhibit complementary advantage regions across benchmarks and task types. $L_{\mathrm{LM}}$ is more favorable for several mathematics- and code-related evaluations. $L_{\mathrm{KD}}$, by contrast, performs better on several general-domain benchmarks---including knowledge-oriented, commonsense, and reading-comprehension evaluations. These complementary patterns motivate a natural question: rather than applying a single objective uniformly across all training data, can objective routing be guided by coarse data-domain labels?

We evaluate a domain-based subset-level routing policy. Following the data mixture described in Section~\ref{subsec:training_data}, we retain the same overall training composition as the single-objective baselines but assign different objectives to different domains. Specifically, we apply $L_{\mathrm{LM}}$ to mathematics and code data, and $L_{\mathrm{KD}}$ with top-$256$ truncation and temperature $\tau=1$ to the remaining general-domain data. Data packing is performed separately within each domain, ensuring that no packed training sequence spans multiple domains. This design makes the objective assignment unambiguous at the sequence level and avoids mixing $L_{\mathrm{LM}}$- and $L_{\mathrm{KD}}$-supervised tokens within the same packed sequence. All other training hyperparameters follow the configuration in Section~\ref{subsec:motivating_observation_experimental_design}. We evaluate the resulting model using the protocol described in Section~\ref{subsec:evaluation_protocol}, and refer to this configuration as \textbf{Domain} throughout the following discussion.

Tables~\ref{tab:domain_routing_benchmark} and~\ref{tab:domain_routing_pass_at_k} report the standard benchmark and Pass@$K$ results, respectively. Across both evaluation settings, domain-level routing consistently improves aggregate performance relative to the single-objective baselines and often matches or exceeds the stronger of the two on individual benchmarks. On the standard benchmark suite, \textbf{Domain} achieves the highest average score among the three variants. On the Pass@$K$ evaluation, where the gap between $L_{\mathrm{LM}}$ and $L_{\mathrm{KD}}$ is more pronounced, \textbf{Domain} substantially outperforms $L_{\mathrm{KD}}$ and is broadly competitive with $L_{\mathrm{LM}}$. These results suggest that coarse domain-level routing can selectively retain the advantages of each objective rather than producing outcomes intermediate between the two single-objective baselines.

\paragraph{Recovering the $L_{\mathrm{LM}}$ advantage where $L_{\mathrm{KD}}$ underperforms.}
The first clear pattern is that \textbf{Domain} largely recovers the $L_{\mathrm{LM}}$ advantage on tasks where $L_{\mathrm{KD}}$ is weak, particularly on high-difficulty reasoning, mathematics, and sampling-intensive evaluations. On MMLU-Pro, $L_{\mathrm{KD}}$ falls noticeably behind $L_{\mathrm{LM}}$, while \textbf{Domain} remains close to $L_{\mathrm{LM}}$. A similar pattern appears on MATH (Minerva), where \textbf{Domain} slightly exceeds $L_{\mathrm{LM}}$ and clearly outperforms $L_{\mathrm{KD}}$. The trend is more pronounced in the Pass@$K$ results. On MATH-500, $L_{\mathrm{KD}}$ is consistently weaker than $L_{\mathrm{LM}}$ across all sampling budgets, whereas \textbf{Domain} remains close to $L_{\mathrm{LM}}$, especially at larger $K$. On AIME 2024 and AIME 2025, $L_{\mathrm{KD}}$ performs poorly, while \textbf{Domain} recovers strong mathematical Pass@$K$ performance and exceeds $L_{\mathrm{LM}}$ in several settings. These results are consistent with the intended effect of the routing policy: applying $L_{\mathrm{LM}}$ to mathematics and code subsets mitigates the degradation observed under distillation-only training on tasks requiring precise token-level supervision.

\paragraph{Retaining the $L_{\mathrm{KD}}$ advantage where $L_{\mathrm{KD}}$ is favorable.}
The second pattern is that \textbf{Domain} retains the benefit of $L_{\mathrm{KD}}$ on several general-domain benchmarks. DROP is the clearest case: $L_{\mathrm{KD}}$ outperforms $L_{\mathrm{LM}}$, and \textbf{Domain} matches the $L_{\mathrm{KD}}$ score, indicating that the routing strategy preserves the distillation advantage on this benchmark. PIQA and TriviaQA show a weaker but consistent version of the same trend, where \textbf{Domain} remains close to $L_{\mathrm{KD}}$ rather than reverting to $L_{\mathrm{LM}}$-only behavior.

\paragraph{Cases where \textbf{Domain} exceeds both single-objective baselines.}
In several cases, \textbf{Domain} exceeds both $L_{\mathrm{LM}}$ and $L_{\mathrm{KD}}$, indicating that the routed model is not always bounded by the two single-objective baselines. On MBPP, \textbf{Domain} achieves a clear improvement over both variants, with a smaller gain also observed on BBH. In the Pass@$K$ evaluation, \textbf{Domain} outperforms both baselines on AIME 2024 and AIME 2025 across several sampling budgets. These results are compatible with the view that $L_{\mathrm{LM}}$ and $L_{\mathrm{KD}}$ provide complementary supervision signals that, when applied selectively, can produce outcomes stronger than either objective alone.

\paragraph{A remaining exception.}
RACE is the main exception to the overall trend: \textbf{Domain} underperforms both $L_{\mathrm{LM}}$ and $L_{\mathrm{KD}}$ on this benchmark, despite the generally favorable behavior of domain-level routing across the rest of the evaluation suite. We currently have no satisfactory explanation for this isolated reversal and leave its investigation to future work.

\subsection{Diagnostic Metrics Across Domains}
\label{subsec:diagnostic_metrics_of_domains}

\begin{figure}[t]
    \centering
    \includegraphics[width=\linewidth]{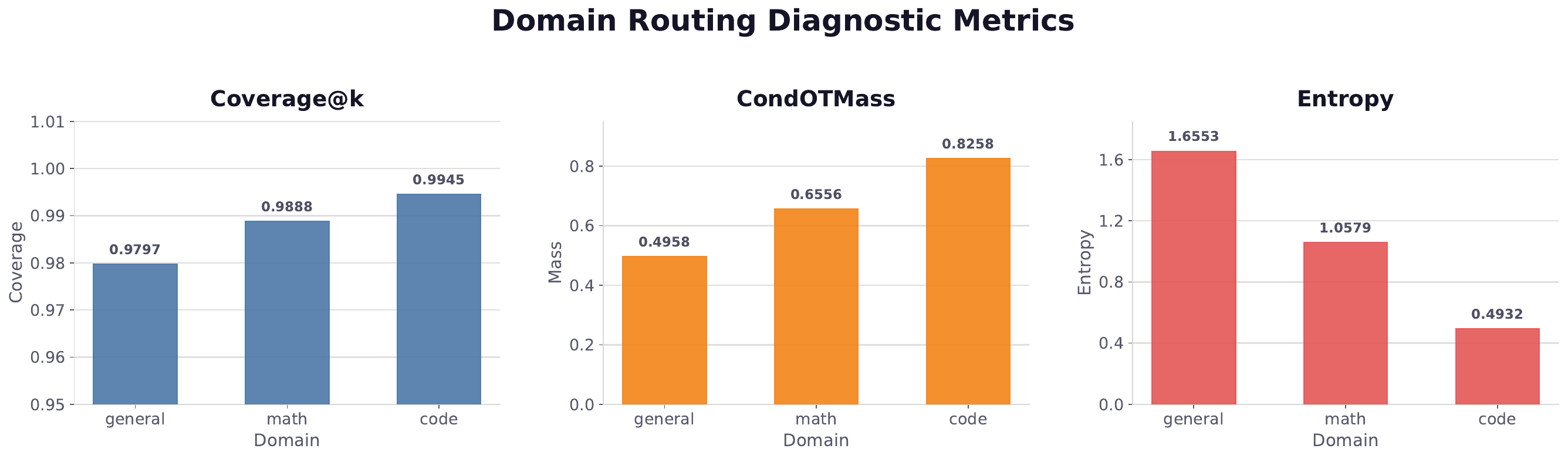}
    \caption{
        Domain-stratified diagnostic metrics under top-$256$ distillation with temperature $\tau=1$.
    }
    \label{fig:domain_routing_diagnostic_metrics}
\end{figure}

The domain-routing results demonstrate that coarse domain labels can serve as a useful prior for objective routing. However, a domain label is inherently a qualitative, high-level annotation: it does not directly characterize the relationship between the observed training token and the teacher target distribution. To examine whether domain labels correlate with more fundamental token-level signals, we follow the analysis protocol of Section~\ref{subsec:effect_of_top_k}. Specifically, we sample 1B tokens from the training corpus and compute the diagnostic metrics defined in Section~\ref{subsec:diagnostic_metrics} separately for each domain, fixing the teacher target to top-$256$ truncation with temperature $\tau=1$.

We focus on three metrics: Coverage@$k$, CondOTMass, and Entropy. Together, these capture complementary aspects of the teacher target distribution. Coverage@$k$ measures whether the observed token falls within the teacher top-$k$ support. CondOTMass measures the probability mass assigned to the observed token by the truncated teacher distribution, conditioned on the observed token being covered. Entropy measures how teacher probability mass is distributed among supported alternatives. The domain-stratified statistics are shown in Figure~\ref{fig:domain_routing_diagnostic_metrics}.

Coverage@$k$ is high across all three domains and is therefore unlikely to be the primary source of domain-level differences under the top-$256$ setting. General-domain data has Coverage@$k$ of $0.9797$, while mathematics and code reach $0.9888$ and $0.9945$, respectively. This indicates that, when the teacher support is sufficiently large, the main distinction across domains lies not in whether the observed token is included in the support.

CondOTMass exhibits a more pronounced domain-level separation. The value increases from $0.4958$ for general-domain data to $0.6556$ for mathematics and $0.8258$ for code. This indicates that, conditioned on coverage, the observed token receives substantially higher teacher probability mass in mathematics and code than in the general domain. CondOTMass is directly related to the within-support component of the $L_{\mathrm{LM}}$--$L_{\mathrm{KD}}$ supervision gap described in Eq.~\eqref{eq:gradient_gap}: $L_{\mathrm{LM}}$ assigns all target mass to the observed token, whereas $L_{\mathrm{KD}}$ assigns only $p_x^{(T,k,\tau)}$ and distributes the remainder over teacher-supported alternatives. Since Coverage@$k$ is consistently high across all domains under top-$256$, domain-level variation in CondOTMass provides a useful proxy for the strength of the observed-token supervision signal. Notably, CondOTMass depends jointly on the observed data token and the teacher distribution, making it a measure of teacher--data alignment rather than a property of the teacher alone.

Entropy shows the opposite trend. General-domain data has the highest entropy ($1.6553$), while mathematics and code have lower values of $1.0579$ and $0.4932$, respectively, indicating that the teacher target distribution is more diffuse on general-domain data and more concentrated on mathematics and code. Unlike CondOTMass, however, Entropy does not directly isolate the observed-token component of the $L_{\mathrm{LM}}$--$L_{\mathrm{KD}}$ gradient gap in Eq.~\eqref{eq:gradient_gap}: it characterizes how teacher probability mass is spread over supported alternatives without reference to whether the observed token is covered or assigned high mass. Entropy therefore reflects teacher-side distributional concentration in isolation, rather than alignment between the teacher target and the observed token.

Taken together, these observations suggest that the informativeness of domain labels is consistent with their correlation with more fundamental token-level statistics. Mathematics and code data tend to exhibit higher CondOTMass and lower Entropy, whereas general-domain data tends to exhibit lower CondOTMass and higher Entropy. Nonetheless, domain labels remain a coarse proxy for these underlying signals. To evaluate more direct routing criteria, we next examine routing policies defined directly in terms of the diagnostic metrics, using OTMass and Entropy as two representative signals.\footnote{The domain-level analysis in Section~\ref{subsec:diagnostic_metrics_of_domains} decomposes OTMass into Coverage@$k$ and CondOTMass to separately characterize coverage and within-support mass variation across domains. For token-level routing, however, CondOTMass is undefined at positions where $x_r \notin K_r^{(T)}$, precluding its direct use as a per-token routing signal. OTMass is well-defined at every position---it equals zero when $x_r \notin K_r^{(T)}$ (Eq.~\eqref{eq:observed_token_mass})---and jointly captures both coverage and within-support mass allocation, making it the appropriate signal for per-token routing decisions.}
\subsection{Token-Level Objective Routing by OTMass and Teacher Entropy}
\label{subsec:token_level_routing}

We evaluate two token-level routing policies that use OTMass and teacher entropy, respectively, as per-token routing signals. Both policies are motivated in part by the objective-selection rationale of~\cite{goyal2026distilled}: $L_{\mathrm{LM}}$ is assigned to tokens for which the teacher distribution is unlikely to provide meaningful supervision beyond the observed token, while $L_{\mathrm{KD}}$ is assigned to tokens for which the teacher may supply useful distributional guidance over alternatives. 

For OTMass-based routing, $L_{\mathrm{KD}}$ with top-$256$ truncation and temperature $\tau=1$ is applied to tokens with low OTMass, and $L_{\mathrm{LM}}$ to the remaining tokens. Based on the OTMass statistics in Section~\ref{subsec:diagnostic_metrics_of_domains}, we select two thresholds corresponding approximately to the 30th and 70th quantiles of the OTMass distribution: $T_M \in \{0.3678, 0.9876\}$. For entropy-based routing, $L_{\mathrm{LM}}$ is applied to tokens with low entropy and $L_{\mathrm{KD}}$ with top-$256$ truncation and temperature $\tau=1$ to tokens with high entropy. Based on the entropy statistics in Section~\ref{subsec:diagnostic_metrics_of_domains}, we select two thresholds corresponding approximately to the 30th and 50th quantiles: $T_E \in \{0.1, 0.5\}$. All other training hyperparameters follow the configuration in Section~\ref{subsec:motivating_observation_experimental_design}, and evaluation follows the protocol in Section~\ref{subsec:evaluation_protocol}. The standard benchmark and Pass@K results are presented in Tables~\ref{tab:otmass_routing_benchmark}, \ref{tab:otmass_routing_pass_at_k}, \ref{tab:entropy_routing_benchmark} and \ref{tab:entropy_routing_pass_at_k}.

\begin{table}[t]
\centering
\caption{Benchmark-level comparison under different OTMass routing strategies.}
\label{tab:otmass_routing_benchmark}
\begin{tabular}{lcccc}
\toprule
\textbf{Benchmark} & \textbf{LM} & \textbf{KD} & \bm{$T_M=0.3678$} & \bm{$T_M=0.9876$} \\
\midrule
PIQA           & 79.16 & 80.63 & 79.65 & \textbf{81.07} \\
HellaSwag      & 79.06 & 79.16 & 78.38 & \textbf{79.31} \\
MMLU           & 68.27 & \textbf{68.83} & 68.30 & 68.64 \\
TriviaQA       & 54.65 & 55.49 & \textbf{56.26} & 55.47 \\
RACE           & \textbf{46.89} & 45.74 & 45.65 & 45.84 \\
DROP           & 51.59 & 53.37 & \textbf{54.45} & 52.74 \\
MMLU-Pro       & \textbf{43.98} & 39.74 & 41.42 & 40.77 \\
BBH            & 69.90 & 69.33 & \textbf{70.50} & 69.54 \\
C-Eval         & \textbf{69.91} & 68.65 & 68.80 & 68.50 \\
CMMLU          & 72.41 & 72.72 & 72.66 & \textbf{72.90} \\
MBPP           & 53.00 & \textbf{54.80} & 53.60 & 52.80 \\
GSM8K          & 70.43 & 70.74 & \textbf{70.89} & 69.52 \\
MATH (Minerva) & \textbf{43.38} & 40.64 & 41.98 & 41.10 \\
\bottomrule
\end{tabular}
\end{table}

\begin{table*}[t]
\centering
\caption{Pass@$K$ comparison under different OTMass routing strategies.}
\label{tab:otmass_routing_pass_at_k}
\resizebox{\textwidth}{!}{
\begin{tabular}{lcccccccccccccccc}
\toprule
\multirow{2}{*}{\textbf{Benchmark}}
& \multicolumn{4}{c}{\textbf{Pass@1}}
& \multicolumn{4}{c}{\textbf{Pass@16}}
& \multicolumn{4}{c}{\textbf{Pass@64}}
& \multicolumn{4}{c}{\textbf{Pass@128}} \\
\cmidrule(lr){2-5}
\cmidrule(lr){6-9}
\cmidrule(lr){10-13}
\cmidrule(lr){14-17}
& \textbf{LM} & \textbf{KD} & \bm{$T_M=0.3678$} & \bm{$T_M=0.9876$}
& \textbf{LM} & \textbf{KD} & \bm{$T_M=0.3678$} & \bm{$T_M=0.9876$}
& \textbf{LM} & \textbf{KD} & \bm{$T_M=0.3678$} & \bm{$T_M=0.9876$}
& \textbf{LM} & \textbf{KD} & \bm{$T_M=0.3678$} & \bm{$T_M=0.9876$} \\
\midrule
MATH-500
& \textbf{13.02} & 3.07 & 4.93 & 4.08
& \textbf{60.75} & 28.53 & 40.41 & 36.64
& \textbf{78.80} & 51.90 & 65.04 & 62.95
& \textbf{84.40} & 63.00 & 74.00 & 73.00 \\
AIME 2024
& \textbf{0.57} & 0.08 & 0.03 & 0.08
& \textbf{5.80} & 1.25 & 0.42 & 1.25
& \textbf{14.17} & 5.00 & 1.67 & 5.00
& \textbf{23.33} & 10.00 & 3.33 & 10.00 \\
AIME 2025
& \textbf{0.23} & 0.00 & 0.13 & 0.16
& \textbf{3.65} & 0.00 & 2.08 & 2.45
& \textbf{13.35} & 0.00 & 8.33 & 9.17
& \textbf{23.33} & 0.00 & 16.67 & 16.67 \\
HumanEval
& \textbf{29.63} & 22.79 & 26.46 & 18.72
& \textbf{77.88} & 72.39 & 74.33 & 66.14
& \textbf{90.25} & 86.25 & 87.15 & 84.12
& \textbf{93.90} & 90.24 & 90.85 & 89.02 \\
\bottomrule
\end{tabular}
}
\end{table*}

\begin{table}[t]
\centering
\caption{Benchmark-level comparison under different entropy routing strategies.}
\label{tab:entropy_routing_benchmark}
\begin{tabular}{lcccc}
\toprule
\textbf{Benchmark} & \textbf{LM} & \textbf{KD} & \bm{$T_E=0.1$} & \bm{$T_E=0.5$} \\
\midrule
PIQA           & 79.16 & 80.63 & \textbf{80.69} & 79.82 \\
HellaSwag      & 79.06 & 79.16 & \textbf{79.39} & 79.01 \\
MMLU           & 68.27 & \textbf{68.83} & 68.42 & 68.53 \\
TriviaQA       & 54.65 & 55.49 & \textbf{55.59} & 55.21 \\
RACE           & \textbf{46.89} & 45.74 & 45.84 & 39.33 \\
DROP           & 51.59 & \textbf{53.37} & 53.22 & 51.37 \\
MMLU-Pro       & \textbf{43.98} & 39.74 & 39.69 & 43.86 \\
BBH            & \textbf{69.90} & 69.33 & 68.19 & 68.70 \\
C-Eval         & 69.91 & 68.65 & \textbf{70.36} & 69.24 \\
CMMLU          & 72.41 & \textbf{72.72} & 72.70 & 72.49 \\
MBPP           & 53.00 & 54.80 & 52.20 & \textbf{55.00} \\
GSM8K          & 70.43 & 70.74 & 68.31 & \textbf{71.34} \\
MATH (Minerva) & \textbf{43.38} & 40.64 & 41.22 & 41.94 \\
\bottomrule
\end{tabular}
\end{table}

\begin{table*}[t]
\centering
\caption{Pass@$K$ comparison under different entropy routing strategies.}
\label{tab:entropy_routing_pass_at_k}
\resizebox{\textwidth}{!}{
\begin{tabular}{lcccccccccccccccc}
\toprule
\multirow{2}{*}{\textbf{Benchmark}}
& \multicolumn{4}{c}{\textbf{Pass@1}}
& \multicolumn{4}{c}{\textbf{Pass@16}}
& \multicolumn{4}{c}{\textbf{Pass@64}}
& \multicolumn{4}{c}{\textbf{Pass@128}} \\
\cmidrule(lr){2-5}
\cmidrule(lr){6-9}
\cmidrule(lr){10-13}
\cmidrule(lr){14-17}
& \textbf{LM} & \textbf{KD} & \bm{$T_E=0.1$} & \bm{$T_E=0.5$}
& \textbf{LM} & \textbf{KD} & \bm{$T_E=0.1$} & \bm{$T_E=0.5$}
& \textbf{LM} & \textbf{KD} & \bm{$T_E=0.1$} & \bm{$T_E=0.5$}
& \textbf{LM} & \textbf{KD} & \bm{$T_E=0.1$} & \bm{$T_E=0.5$} \\
\midrule
MATH-500  & \textbf{13.02} & 3.07 & 3.46 & 3.14 & \textbf{60.75} & 28.53 & 32.71 & 31.22 & \textbf{78.80} & 51.90 & 59.10 & 58.30 & \textbf{84.40} & 63.00 & 70.40 & 69.40 \\
AIME 2024 & \textbf{0.57}  & 0.08 & 0.08 & 0.10 & \textbf{5.80}  & 1.25  & 1.25  & 1.67  & \textbf{14.17} & 5.00  & 5.00  & 6.67  & \textbf{23.33} & 10.00 & 10.00 & 13.33 \\
AIME 2025 & \textbf{0.23}  & 0.00 & 0.08 & 0.08 & \textbf{3.65}  & 0.00  & 1.20  & 1.20  & \textbf{13.35} & 0.00  & 4.17  & 4.17  & \textbf{23.33} & 0.00  & 6.67  & 6.67  \\
HumanEval & \textbf{29.63} & 22.79 & 24.31 & 23.81 & \textbf{77.88} & 72.39 & 72.87 & 73.01 & \textbf{90.25} & 86.25 & 87.35 & 87.15 & \textbf{93.90} & 90.24 & 92.07 & 90.85 \\
\bottomrule
\end{tabular}
}
\end{table*}

\paragraph{Token-level routing does not provide a consistent overall gain over $L_{\mathrm{LM}}$ or $L_{\mathrm{KD}}$.}
The results for OTMass routing are mixed. On DROP and TriviaQA, where $L_{\mathrm{KD}}$ is stronger than $L_{\mathrm{LM}}$, the $T_M=0.3678$ variant exceeds both single-objective baselines: it achieves $54.45$ on DROP (vs.\ $53.37$ for $L_{\mathrm{KD}}$ and $51.59$ for $L_{\mathrm{LM}}$) and $56.26$ on TriviaQA (vs.\ $55.49$ and $54.65$). On BBH, the same variant also surpasses both baselines, reaching $70.50$ against $69.33$ for $L_{\mathrm{KD}}$ and $69.90$ for $L_{\mathrm{LM}}$. These gains are consistent with the view that OTMass captures a useful per-token routing signal, though benchmark-level aggregates alone do not identify the specific token subsets responsible.

These improvements are not systematic, however. On benchmarks where $L_{\mathrm{LM}}$ is clearly stronger, OTMass routing only partially recovers the $L_{\mathrm{LM}}$ advantage: on MMLU-Pro, the two OTMass variants achieve $41.42$ and $40.77$ against $43.98$ for $L_{\mathrm{LM}}$; on MATH (Minerva), they achieve $41.98$ and $41.10$ against $43.38$. The limitation is most pronounced in pass@$K$ evaluation: on MATH-500 Pass@128, the best OTMass variant reaches $74.00$, a partial recovery from $L_{\mathrm{KD}}$ ($63.00$) but still well short of $L_{\mathrm{LM}}$ ($84.40$). In aggregate, OTMass routing mitigates the degradation associated with $L_{\mathrm{KD}}$ but does not consistently restore the performance of $L_{\mathrm{LM}}$ on high-difficulty reasoning and generation tasks.

Entropy-based routing shows a similar lack of consistent dominance. The $T_E=0.5$ variant produces competitive results on several benchmarks---nearly matching $L_{\mathrm{LM}}$ on MMLU-Pro ($43.86$ vs.\ $43.98$) and marginally exceeding both baselines on GSM8K ($71.34$ vs.\ $70.43$ and $70.74$) and MBPP ($55.00$ vs.\ $54.80$)---and improves AIME 2024 Pass@128 from $10.00$ under $L_{\mathrm{KD}}$ to $13.33$. However, these gains do not form a consistent pattern, and entropy routing can produce marked regressions. The most notable case is RACE: $L_{\mathrm{LM}}$ obtains $46.89$ and $L_{\mathrm{KD}}$ obtains $45.74$, whereas the $T_E=0.5$ variant drops to $39.33$. On BBH, both entropy variants ($68.19$ and $68.70$) fall below $L_{\mathrm{LM}}$ ($69.90$), $L_{\mathrm{KD}}$ ($69.33$), and the OTMass routing results. These observations suggest that teacher entropy alone is an insufficient and potentially unreliable signal for per-token objective selection.

\paragraph{OTMass routing is more stable than entropy routing.}
Although neither token-level routing strategy consistently outperforms both single-objective baselines, OTMass routing is more stable across the benchmarks considered here, particularly in avoiding large regressions. On RACE, OTMass routing remains close to $L_{\mathrm{KD}}$, with scores of $45.65$ and $45.84$, whereas entropy routing with $T_E=0.5$ drops to $39.33$. On BBH, OTMass routing obtains $70.50$ and $69.54$, compared with $68.19$ and $68.70$ for entropy routing. On GSM8K, OTMass routing varies from $69.52$ to $70.89$, against a wider spread of $68.31$ to $71.34$ for entropy routing. In pass@$K$ evaluation, OTMass routing also provides a more predictable interpolation between the two single-objective baselines than entropy routing does.

This difference is consistent with the properties of the two metrics.
OTMass depends jointly on the teacher distribution and the observed training token: it measures the probability mass assigned by the teacher to the observed token under top-$k$ truncation and is zero when the observed token falls outside the support. It is thus closely connected to the observed-token component of the $L_{\mathrm{LM}}$--$L_{\mathrm{KD}}$ gradient discrepancy described in Eq.~\eqref{eq:gradient_gap}, and may be expected to capture teacher--data alignment more directly than teacher-side statistics alone. Entropy, by contrast, characterizes only the concentration of the teacher distribution, independently of the observed token. A low-entropy teacher distribution indicates that the teacher assigns high confidence to some token, but that token need not be the observed one. As entropy does not condition on the observed training token, it cannot in general distinguish between a teacher that agrees with the data and one that confidently favors an alternative; this may contribute to entropy-based routing being more sensitive to threshold choice and more susceptible to abrupt performance regressions.

\paragraph{Domain-level subset routing appears to outperform these token-level routing strategies.}
Compared with both token-level policies, domain-based subset-level routing tends to yield a more favorable and more consistent improvement pattern. On the standard benchmark suite, domain routing recovers the $L_{\mathrm{LM}}$ advantage on tasks where $L_{\mathrm{KD}}$ underperforms while preserving the $L_{\mathrm{KD}}$ advantage where it is favorable. On MMLU-Pro, domain routing achieves $43.72$---close to $L_{\mathrm{LM}}$ ($43.98$) and well above the best OTMass variant ($41.42$). On MATH (Minerva), domain routing reaches $43.58$, slightly exceeding $L_{\mathrm{LM}}$ ($43.38$) and outperforming all token-level routing variants. On MBPP, domain routing achieves $58.40$, surpassing $L_{\mathrm{LM}}$ ($53.00$), $L_{\mathrm{KD}}$ ($54.80$), and the best token-level results.

The advantage of domain routing is more pronounced in pass@$K$ evaluation. On MATH-500 Pass@128, domain routing reaches $83.20$, far closer to $L_{\mathrm{LM}}$ ($84.40$) than the best OTMass routing result ($74.00$). On AIME 2024 Pass@128, domain routing achieves $33.33$, exceeding $L_{\mathrm{LM}}$ ($23.33$), $L_{\mathrm{KD}}$ ($10.00$), and the best token-level result ($13.33$); a similar pattern holds on AIME 2025, where domain routing matches $L_{\mathrm{LM}}$ ($23.33$) while the best token-level variant reaches only $16.67$. On HumanEval at Pass@128, domain routing obtains $92.68$, compared with $92.07$ for the best entropy routing variant and $90.85$ for the best OTMass variant. These results indicate that, within the current experimental setup, coarse domain information provides a more effective basis for objective routing than the per-token statistics examined here.

One possible explanation for this gap is that token-level routing may be more sensitive to context contamination introduced by the pre-training data format. In large-scale pre-training, multiple samples are typically packed into fixed-length sequences, and training is often performed without cross-sample attention masks for efficiency. As a result, teacher logits at any position may reflect not only the local sample content but also adjacent packed samples, potentially perturbing fine-grained per-token quantities such as OTMass and entropy and introducing noise into the routing decisions. Domain-level routing, operating at a coarser granularity, is likely less sensitive to such local perturbations. We leave a more systematic investigation of this packed-context effect and its implications for token-level objective routing to future work.
\subsection{Summary of Findings}

The experimental results provide a qualified but substantive answer to RQ2. Training objectives can be routed adaptively across the training distribution using statistics derived from the training data, and such adaptive routing offers a more targeted mechanism for managing capability trade-offs than a fixed global objective. The effectiveness of this approach, however, appears to depend largely on the routing signal: increasing routing granularity alone does not appear sufficient, and the routing criterion may need to be reasonably aligned with the capability differences induced by the two objectives.

The clearest evidence for adaptive routing comes from domain-based subset-level routing. Assigning $L_{\mathrm{LM}}$ to mathematics and code subsets and $L_{\mathrm{KD}}$ to the remaining general-domain data yields consistent aggregate improvements over both single-objective baselines, and in several cases---including MBPP and the AIME benchmarks---the routed model exceeds both $L_{\mathrm{LM}}$ and $L_{\mathrm{KD}}$ individually. The routing strategy largely recovers the advantage of $L_{\mathrm{LM}}$ on tasks where distillation underperforms, particularly on high-difficulty reasoning and sampling-intensive evaluations, while preserving the advantage of $L_{\mathrm{KD}}$ on knowledge-oriented and commonsense benchmarks where teacher-distribution supervision is beneficial. These results indicate that coarse domain structure provides a reliable basis for objective routing, at least for broad capability dimensions such as mathematical reasoning, coding, and general-domain language understanding, and suggest that selectively combining complementary supervision signals can produce outcomes stronger than either objective alone.

Between the two token-level signals, OTMass appears to be the more reliable routing criterion. This is broadly consistent with its interpretation as a joint teacher--data alignment statistic: OTMass directly measures the teacher probability mass assigned to the observed training token under top-$k$ truncation, and is thus closely connected to the observed-token component of the $L_{\mathrm{LM}}$--$L_{\mathrm{KD}}$ gradient discrepancy characterized in Eq.~\eqref{eq:gradient_gap}. Teacher entropy, by contrast, reflects only the concentration of the teacher distribution independently of the observed token; it cannot distinguish between a teacher that agrees with the data and one that confidently favors an alternative, which may make it a less principled and more threshold-sensitive routing criterion.

These findings point to several tentative conclusions about the design of effective routing signals. First, signals that jointly reflect the teacher distribution and its alignment with the observed training token may be more informative for objective routing than teacher-only uncertainty measures. Second, at token-level granularity, routing signals may be susceptible to noise introduced by the packed-sequence training format, in which teacher logits at a given position can be influenced by adjacent samples packed into the same sequence; operating at coarser granularities may average over such local variability and yield more stable routing decisions. Third, and most broadly, the advantage of domain-level routing over the token-level policies evaluated here suggests that the effectiveness of adaptive routing need not increase monotonically with routing granularity. Identifying more principled routing criteria---including sequence-level, capability-aware, and learned routing signals---remains an important direction for future work.

\section{Conclusion}

\paragraph{Summary.}
This paper examines continued pre-training via off-policy distillation through the joint lens of token-level supervision, KD parameterization, and data heterogeneity. We begin by establishing that $L_{\mathrm{LM}}$ and $L_{\mathrm{KD}}$ do not differ merely in degree but in kind: under otherwise identical training conditions, the two objectives induce systematically distinct capability profiles. $L_{\mathrm{LM}}$ is consistently stronger on high-difficulty reasoning, mathematical problem solving, and knowledge-intensive evaluations---with performance gaps that widen substantially under large-budget Pass@$K$ evaluation---while $L_{\mathrm{KD}}$ is more favorable on commonsense plausibility, factual retrieval, reading comprehension, and structured program synthesis. To explain this divergence, we analyze the two objectives at the gradient level and identify a fundamental tension between two complementary supervision signals: \emph{direct observed-token reinforcement}, which is the exclusive mode of $L_{\mathrm{LM}}$, and \emph{teacher-supported alternative supervision}, which $L_{\mathrm{KD}}$ introduces by distributing target mass across the teacher top-$k$ support. We introduce a suite of diagnostic metrics---Coverage@$k$, OTRank, OTMass, CondOTMass, and teacher-support entropy---to quantify this balance at each training position. Controlled sweeps over the two principal KD degrees of freedom show that the support size $k$ and distillation temperature $\tau$ operate through distinct mechanisms: $k$ governs a coverage--sharpness trade-off in which expanding the support reduces observed-token exclusion at the cost of a progressively more diffuse target, while $\tau$ controls within-support probability allocation without altering support membership. The resulting downstream effects are qualitatively distinct and task-dependent, consistent with the view that different tasks favor different positions along the spectrum between the two complementary supervision signals.

We then investigate whether objective routing can be made adaptive to the training data. A domain-based routing policy---applying $L_{\mathrm{LM}}$ to mathematics and code subsets and $L_{\mathrm{KD}}$ to general-domain data---yields consistent aggregate improvements over both single-objective baselines, recovering the $L_{\mathrm{LM}}$ advantage on reasoning-intensive and sampling-intensive evaluations while preserving the $L_{\mathrm{KD}}$ advantage on knowledge-oriented and commonsense benchmarks; in several cases, including MBPP and the AIME benchmarks, the routed model exceeds both baselines simultaneously. Analysis of the diagnostic metrics stratified by domain reveals that this effectiveness does not appear to be incidental: mathematics and code data exhibit substantially higher OTMass and lower teacher-support entropy than general-domain data, indicating that domain labels correlate with more fundamental differences in teacher--data alignment. Finer-grained token-level routing via OTMass and teacher entropy, however, does not consistently match the stronger single-objective baseline and falls well short of domain-level routing on high-difficulty reasoning evaluations. Between the two token-level signals, OTMass appears to be the more reliable criterion, plausibly in that it jointly reflects the teacher distribution and its alignment with the observed training token, whereas teacher entropy characterizes only teacher-side distributional concentration and cannot distinguish a teacher that agrees with the data from one that confidently predicts an alternative. Taken together, these findings suggest that routing effectiveness need not increase monotonically with routing granularity, and that principled objective routing may instead call for signals grounded in teacher--data alignment rather than teacher-side uncertainty alone.

\paragraph{Limitations.}
This work has several limitations.
First, the gap between domain-level and token-level routing also remains incompletely explained. Our experiments suggest that domain-based subset routing tends to outperform per-token policies based on OTMass or teacher entropy, suggesting that finer routing granularity is not necessarily advantageous in itself. A plausible contributing factor is the sequence-packing format common in large-scale pre-training: when multiple documents are packed into a single fixed-length sequence without cross-sample attention masking, teacher logits at a given position may reflect surrounding packed content rather than the local document context alone, perturbing token-level diagnostic statistics and weakening their alignment with the intended routing criterion. Whether this contamination fully accounts for the observed gap---and whether token-level routing with cleaner document boundaries or attention masks can recover stronger performance---remains an open question.

Second, our capability characterization is grounded in benchmark performance, which is an imperfect proxy for isolated model abilities. Individual benchmarks typically conflate multiple factors---factual recall, multi-step reasoning, reading comprehension, numerical manipulation, and decoding sensitivity---so the performance profiles reported here should be interpreted as empirical regularities at the benchmark level rather than as clean decompositions of latent capabilities. Establishing sharper causal links between training-objective choices and specific capability dimensions would require evaluation protocols with explicitly annotated skill factors and controlled data-to-capability mappings.

Finally, due to computational and experimental budget constraints, all experiments are conducted under a continued pre-training setup, in which the student is warm-started from a checkpoint derived via a pruning--distillation scaling ladder, rather than trained from scratch; in addition, we do not evaluate our method across different model families. Whether the qualitative patterns reported here---particularly the optimal ranges of $k$ and $\tau$ and the relative advantage of domain-level over token-level routing---transfer to from-scratch pre-training regimes or to other model families remains an open question.

\paragraph{Future Work.}
A natural extension of this work is to develop a principled framework for adaptive objective routing in off-policy distillation, organized around three components: the routing signal, the routing objective, and the routing rule. On the signal side, the central questions are what information should govern objective assignment and at what granularity routing decisions should be made. Routing can operate at the level of domains, documents, sequences, spans, or individual tokens, and finer granularity trades stability for adaptivity while also increasing susceptibility to local noise---including the packed-sequence contamination discussed in Section~\ref{subsec:token_level_routing}. Candidate signals include domain labels, observed-token probability mass, teacher--data alignment metrics, and gradient-based diagnostics.

On the objective side, the design space extends beyond the binary choice between $L_{\mathrm{LM}}$ and $L_{\mathrm{KD}}$ explored here. The KD family spans a continuous range of supervision signals parameterized by the top-$k$ support size, distillation temperature, LM--KD interpolation weight, and the choice of divergence measure, each of which shifts the balance between direct observed-token reinforcement and teacher-supported alternative supervision in a distinct way. A more general routing framework could assign not only whether to apply distillation, but also which configuration within the KD family is most appropriate for a given data region---treating sparse KD as a structured family of supervision signals rather than a monolithic objective, and enabling more targeted capability shaping across heterogeneous training data.

On the rule side, the threshold-based policies evaluated in this work are interpretable and computationally inexpensive, but may be suboptimal when multiple signals interact or when capability trade-offs are task-specific in ways that fixed thresholds cannot capture. More flexible alternatives, such as learned routing policies that map diagnostic features to objective assignments, raise further questions about router supervision, training stability, computational overhead, and generalization across model scales and data distributions. Taken together, these directions point toward a broader reconception of off-policy distillation: rather than a global objective to be tuned empirically, it is more productively framed as a structured supervision design problem whose central challenge is to characterize and exploit the interplay among data heterogeneity, training objective, and model capability.

\bibliographystyle{unsrtnat}
\bibliography{references}

\end{document}